\documentclass[a4paper,fleqn]{cas-dc}

\usepackage{cleveref}
\usepackage[numbers]{natbib}
\usepackage{amssymb}
\usepackage{amsmath,amsfonts}
\usepackage{algorithmic}
\usepackage{algorithm}

\usepackage{array}
\usepackage[caption=false,font=footnotesize,labelfont=sf,textfont=sf]{subfig}
\usepackage{textcomp}
\usepackage{stfloats}
\usepackage{url}
\usepackage{verbatim}
\usepackage{graphicx}
\def\BibTeX{{\rm B\kern-.05em{\sc i\kern-.025em b}\kern-.08em
    T\kern-.1667em\lower.7ex\hbox{E}\kern-.125emX}}
\usepackage{color}
\usepackage{balance}
\usepackage{pifont}
\usepackage{tabularx}
\usepackage{booktabs}
\usepackage{multirow}
\usepackage{etoolbox}
\usepackage{bbm}
\usepackage{bm}
\usepackage{makecell}
\usepackage{dsfont} 

\makeatletter
\patchcmd{\@makecaption}
  {\scshape}
  {}
  {}
  {}
\makeatletter
\patchcmd{\@makecaption}
  {\\}
  {.\ }
  {}
  {}

\makeatother

\def\tsc#1{\csdef{#1}{\textsc{\lowercase{#1}}\xspace}}
\tsc{WGM}
\tsc{QE}

\usepackage{caption}
\hypersetup{
    colorlinks=true,
    citecolor=green,
    linkcolor=red,
    urlcolor=red
}

\begin{document}
\let\WriteBookmarks\relax
\def\floatpagepagefraction{1}
\def\textpagefraction{.001}
\let\printorcid\relax 
\shorttitle{DMSD-CDFSAR}    
\shortauthors{Fei Guo et al.}

\title[mode = title]{DMSD-CDFSAR: Distillation from Mixed-Source Domain for Cross-Domain Few-shot Action Recognition}  

\author[1]{Fei Guo}[type=editor,
    auid=000]
\ead{co.fly@stu.xjut.edu.cn} 

\author[1]{YiKang Wang}
\ead{funnyQ@stu.xjtu.edu.cn}
 
\author[1]{Han Qi}
\ead{qihan19@stu.xjtu.edu.cn}

\author[1]{Li Zhu}
\ead{zhuli@xjtu.edu.cn} 
\cormark[1]

\author[2]{Jing Sun}
\ead{jing.sun1@siat.ac.cn}

\address[1]{School of Software Engineering, Xi'an Jiaotong University, China}
\address[2]{Shenzhen Institutes of Advanced Technology, Chinese Academy of Sciences, China}

\cortext[1]{Corresponding author} 


\makeatletter\def\Hy@Warning#1{}\makeatother
\maketitle
\begin{abstract}
Few-shot action recognition is an emerging field in computer vision, primarily focused on meta-learning within the same domain. However, challenges arise in real-world scenario deployment, as gathering extensive labeled data within a specific domain is laborious and time-intensive. Thus, attention shifts towards cross-domain few-shot action recognition, requiring the model to generalize across domains with significant deviations.
Therefore, we propose a novel approach, ``Distillation from Mixed-Source Domain", tailored to address this conundrum. Our method strategically integrates insights from both labeled data of the source domain and unlabeled data of the target domain during the training. 
The ResNet18 is used as the backbone to extract spatial features from the source and target domains. 
We design two branches for meta-training: the original-source and the mixed-source branches.
In the first branch, a Domain Temporal Encoder is employed to capture temporal features for both the source and target domains. Additionally, a Domain Temporal Decoder is employed to reconstruct all extracted features.
In the other branch, a Domain Mixed Encoder is used to handle labeled source domain data and unlabeled target domain data, generating mixed-source domain features.
We incorporate a pre-training stage before meta-training, featuring a network architecture similar to that of the first branch.
Lastly, we introduce a dual distillation mechanism to refine the classification probabilities of source domain features, aligning them with those of mixed-source domain features. 
This iterative process enriches the insights of the original-source branch with knowledge from the mixed-source branch, thereby enhancing the model's generalization capabilities.
This comprehensive framework enables robust few-shot action recognition across diverse domains, demonstrating the effectiveness of our approach in addressing real-world challenges.
Our code is available at URL: \url{https://xxxx/xxxx/xxxx.git}
\end{abstract}

\section{Introduction} \label{introduction}
\begin{figure}[htbp]
    \centering
    \includegraphics[width=6.2cm]{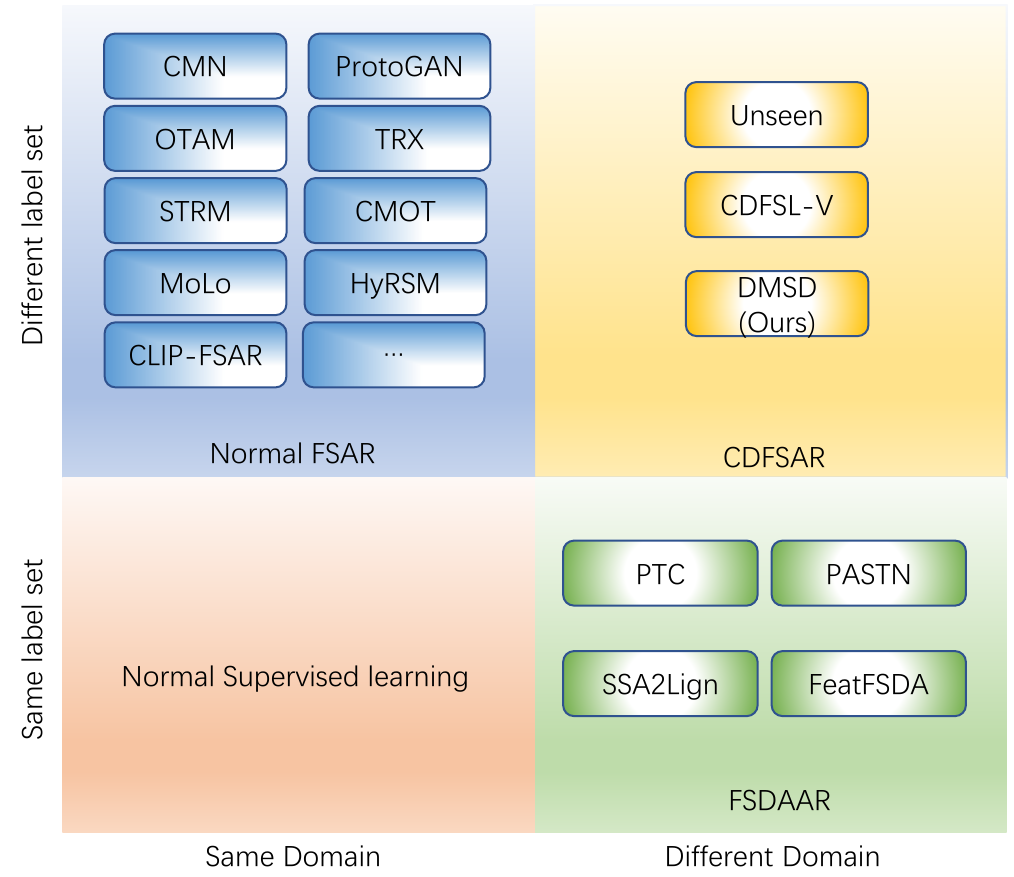}
    \caption{The differences between FASR, CDFSAR, and FSDASR are related to domain and label settings. Representative works corresponding to the issues in each area are exemplified in different colored regions.}
 \label{fig:1}
\end{figure}

Few-shot action recognition (FSAR) involves training a model on a substantial number of labeled videos during the meta-training stage, aiming to generalize its performance to the meta-testing stage, where it must classify unlabeled videos based on only a few labeled videos. Although significant advancements have been made in few-shot action recognition, it is commonly assumed that the data in both meta-training and meta-testing stages come from the same domain. However, this assumption is often untenable in real-world production scenarios. 
Therefore, the importance of Cross-Domain Few-Shot Action Recognition (CDFSAR) is becoming increasingly evident. 
To the best of our knowledge, while considerable progress has been made in Cross-Domain Few-Shot Learning (CDFSL) within the realm of image classification, the exploration of CDFSAR remains relatively underexplored.
Before studying CDFSAR, some researchers focused on Few-shot Domain Adaptation Action Recognition (FSDAAR), similar to CDFSAR but not the same. 
These include:
PTC\cite{gao2020pairwise-domainAdaptionAction},
PASTN\cite{gao2020pairwise-domainAdaptionAction-R2},
FeatFSDA \cite{peng2023featfsda-domainAdaptionAction},
SSA2lign \cite{xu2023augmenting-domainAdaptionAction-A2S},
FS-ADA \cite{li2021supervised-radarActionDomainAdaption}.
These five works primarily focus on domain adaptation. Tasks in the source and target domains are identical, meaning they share the same label space. Additionally, there are only a few labeled data from the target domain. The differences and connections between CDFSAR and FSDAAR are shown in \Cref{fig:1}.

CDFSL addresses the issue of reduced FSL performance resulting from differences in data distribution between the source and target domains. It confronts the amalgamation of challenges inherent in transfer learning and few-shot learning, including domain gaps, class shifts between datasets, and the limited availability of target domain data. Consequently, CDFSL presents a particularly demanding task.
Similarly, CDFSAR shares a common objective with CDFSL. However, CDFSAR extends this challenge to video action recognition, which introduces an additional temporal dimension. This temporal dimension adds complexity to the task, requiring models to discern not only spatial semantics but also understand temporal dynamics, thereby increasing the complexity of the problem.
CDFSAR-SEEN \cite{wang2023cross-crossDomainAction} introduces a supervised temporal network for temporal modeling and incorporates self-supervised learning to acquire more transferable representations during training. During testing, it applies meta-learning to CDFSAR, enabling direct distance comparison without the need for fine-tuning.
CDFSL-V \cite{samarasinghe2023cdfsl-crossDomainAction-cdfsl-v} leverages self-supervised learning to pre-train the VideoMAE backbone and employs distillation and curriculum learning to balance information from both the source and target domains. This approach involves utilizing pseudo labels for the target during training and fine-tuning the network using a few labeled videos from the target domain.
A notable similarity between the two approaches in CDFSAR is their utilization of unlabeled data from the target domain and source domain during the training stage. 
Both approaches have demonstrated that incorporating unlabeled data from the target domain positively impacts CDFSAR performance. Our work will also adhere to this paradigm.

While existing methods have made strides in addressing the challenges of Cross-Domain Few-Shot Action Recognition (CDFSAR), there are still areas with potential for improvement.
CDFSL-V utilizes fine-tuning with labeled videos from the target domain, which can risk overfitting and lead to forgetting previously learned transferable knowledge. 
SEEN relies solely on self-supervised learning with unlabeled target domain data, potentially limiting its ability to exploit the available information fully. Moreover, discrepancies between training and testing paradigms may hinder the optimal transfer of feature extraction.

To address these challenges, we introduce Distillation from Mixed-Source Domain (DMSD), a holistic approach integrating meta-learning with supervised learning. Utilizing ResNet18 as the backbone, our method encompasses pre-training and meta-training stages.
\textbf{In the pre-training stage}, in this stage, we employ the same modules as the original-source branch in the upcoming meta-training, relying solely on supervised learning and feature reconstruction.
\textbf{In the meta-training stage}, we use the metric-based method to help mitigate overfitting issues stemming from fine-tuning with limited labeled videos.
Inspired by CLFS \cite{chen2019closer-cdfs-closerlook}, which underscores the limitations of existing metric-based few-shot methods, our DMSD tries to combine supervised learning with meta-learning for CDFSAR.  
Our DMSD also tries to maximize the utilization of training data through feature transformation and fusion perspectives. 
In this stage, our DMSD comprises two branches.
\textbf{Original-Source Branch}: 
this branch uses a Domain Temporal Encoder to extract features from both domains. It also uses a Domain Temporal Decoder with Cycle Consistency Loss to reconstruct the target domain's spatial features. This process encodes shared representations, aiding in learning commonalities for cross-domain recognition. In addition to supervised learning, it employs a meta-learning method for the source domain. The branch is designed to enhance the model's feature extraction and metric comparison adaptability.
\textbf{Mixed-Source Branch}: 
this branch contains an Instance Center Calibration (ICC) that leverages target domain sample correlations to determine a refined target domain center. Utilizing this center as the $Query$ and the source domain data as $Key$ and $Value$, it enhances the source data via a cross-transformer known as the Domain Mixer Encoder.
Then, we can get generalized data containing information in the target domain. 
We also use both supervised learning and meta-learning for the generalized data in this branch.
Additionally, we utilize the Dual Distillation between the two branches, with the Domain Mixer Encoder serving as the teacher and the Domain Temporal Encoder as the student. This process compels the original-source branch to extract knowledge from the mixed-source branch, enhancing its generalization ability.
The Domain Mixer Encoder in the mixed-source branch is updated to an exponential moving average (EMA) of the Domain Temporal Encoder in the original-source branch, generating robust classification predictions and improving generalization performance.
During the inference stage, we use the original-source branch for cross-domain evaluation, ensuring the utilization of learned cross-domain knowledge.

Our contributions are summarized as follows:
\begin{itemize}
\item We integrate supervised learning with meta-learning, ensuring that the model possesses the capability for meta-learning inference while maintaining generalization ability.
\item We leverage the Domain Temporal Encoder and Decoder in the original-source branch to perform feature encoding and reconstruction, ensuring the correctness of feature extraction and promoting domain sharing across the source and target domains.
\item We utilize the Domain Mixer Encoder in the mixed-source branch to handle labeled source domain data and unlabeled target domain data for generating mixed-source domain features.
\item Our approach employs dual distillation from the mixed-source branch to the original-source branch. One distillation targets meta-learning, and the other supports supervised learning, enabling the original-source branch to acquire knowledge from the mixed-source branch.
\item Extensive experiments across various benchmarks, including HMDB51, UCF101, Kinetics, Diving48, RareAct and SSV2, validate that our approach is competitive with the state-of-the-art methods. 
\end{itemize}

\section{Related Work}
\subsection{Few-shot learning}
Few-shot learning, characterized by using significantly fewer samples than traditional deep learning, has garnered considerable attention for its ability to achieve comparable or superior performance. Over recent years, researchers have categorized few-shot learning into three main approaches.
Metric-based methods classify samples by measuring their similarity, commonly using prototype and matching networks
\cite{simon2020adaptive-few_shot_learning1, snell2017prototypical_few_shot_learning2,sung2018learning-relation-networkfew_shot_learning3, melekhov2016-siamese_few_shot_learning4, vinyals2016matching—few_shot_learning5, liu2018learning-TNP-few_shot_learning6,2023Constructing-expert-3,LIM2024122173-expert-5}.
Model-based methods utilize specific models to adapt to new tasks, enabling few-shot learning quickly
 \cite{finn2017model-MAML-few-shot-model1, rusu2018-metaL-few-shot-model3, 9999670-generalization-of-MAML,kbs-zheng2023detach-meta-transfer, 2022Few-expert-1, WANG2024121971-expert-6}.
Data-augmentation-based methods generate more samples through data augmentation techniques such as image rotation and cropping to improve the model's generalization ability
\cite{ratner2017learning-few-augmentation2,chen1804semantic-few-augmentation3,perez2017effectiveness-few-augmentation1, pahde2021multimodal-few-augmentation4, 10035001-generation-few-shot-learning}.
For instance, 
Prototypical Network \cite{snell2017prototypical_few_shot_learning2} employs a methodology that quantifies the Euclidean distances between prototypes and queries.
Siamese Network \cite{melekhov2016-siamese_few_shot_learning4} calculates image similarity for pairs, and MAML \cite{finn2017model-MAML-few-shot-model1} adopts a meta-learning strategy to facilitate rapid adaptation to new tasks by initializing network parameters effectively. 
Multimodal prototypical network \cite{pahde2021multimodal-few-augmentation4} tackles data scarcity by generating extra images from text descriptions.

\subsection{Cross-domain Few-shot learning}
Traditionally, few-shot learning relies on source and target datasets drawn from the same domain. However, in the context of Cross-Domain Few-Shot Learning, these datasets originate from distinct domains. 
CLFC \cite{chen2019closer-cdfs-closerlook} discovered that conventional few-shot learning methods falter in addressing domain shifts, performing even worse than baseline methods in cross-domain scenarios. 
To address the problem, CDFSL-FWT \cite{tseng2020cross-cdfsl-fwt} incorporates a learnable feature-wise transformation layer within a theta-learning-based feature encoder to mitigate domain shifts across different domains. While this work marks the inception of CDFSL in computer vision, it is constrained by its effectiveness primarily in scenarios with minimal domain gaps. 
BSCDFSL \cite{guo2020broader} compares images from diverse categories beyond natural images, highlighting that meta-learning-based few-shot learning methods underperform compared to simple fine-tuning methods. Furthermore, recent research efforts leverage unlabeled data from the target domain to learn domain-specific representations and enhance performance. 
For instance, STARTUP \cite{phoo2020self-startup} utilizes a fixed pre-training model to generate pseudo labels for unlabeled samples, subsequently training the network with labeled base data and pseudo-labeled target data. 
Similarly, CDFS-UNLABELLED \cite{yao2021cross-cdfs-unlabelled} incorporates unlabeled target domain data to bridge the source-target gap, employing a self-supervised learning approach to maximize knowledge utilization from both labeled and unlabeled training sets. 
DDN \cite{islam2021dynamic-cdfs-dynamic-distillation} employs dynamic distillation to train a feature extractor with augmented unlabeled target data and labeled source data, enhancing few-shot learning performance on the target domain. 
Additionally, LDP-net \cite{zhou2023revisiting-xigongda} introduces a Local Global Distillation Prototype that features a dual branch to classify query images and random local crops separately, followed by knowledge distillation between branches to enhance class consistency. 
ASC \cite{lu2023adaptive-cdfs-adaptiveSemanticConsistency} proposes a new Adaptive Semantic Consistency framework, which adaptively preserves target-related information in the source domain.

\subsection{Few-shot action recognition}
Recently, few-shot action recognition has become a prominent research focus. It aims to solve the unrealistic problem of getting numerous annotated video samples. There are several important works:
CMN\cite{zhu2018compound} and CMN-J\cite{zhu2020label} encode the video representation through the use of significant embeddings, with keys and values readily stored in memory for easy updates.
ProtoGAN \cite{kumar2019protogan} leverages a conditional generative adversarial framework, employing specific semantics to produce new class instances for application in few-shot learning.
AMeFu-Net\cite{fu2020depth} is also related to data augmentation depending on the depth features.
TARN \cite{bishay2019tarn} pioneers the integration of attention mechanisms for temporal synchronization at the frame level.
OTAM \cite{cao2020few-otam} generates a distance matrix based on the DTW \cite{muller2007dynamic} approach, ensuring a precise alignment.
TRX \cite{perrett2021temporal-trx} employs a Cross-Transformer to process video frame tuples, subsequently achieving feature embedding alignment of these tuples.
STRM \cite{thatipelli2022spatio-strm} adds some pre-processing related to spatial attention of feature enrichment for TRX. 
TSA-MLT \cite{guo2024task} performs irrelevant frame filtering and effectively utilizes the alignment of tuples at different levels in TRX.
ARN \cite{zhang2020few-ARN} focuses on robust similarity metrics and spatial-temporal representation of short and long-range action patterns through permutation-invariant pooling and attention.
HCR\cite{li2022hierarchical-HCR}  leverages the Wasserstein metric for aligning subsequences, providing a method that captures the intrinsic structure and meaning within video sequences.
CMOT \cite{lu2021few-cmot} incorporates Optimal Transport theory for video content comparison, emphasizing not just the sequence alignment but also delving into the semantic aspects of the video.
AMFAR \cite{wanyan2023active} uses bidirectional distillation to capture differentiated task-specific knowledge from reliable modality to improve the representation of unreliable one.
MoLo\cite{wang2023molo} crafts a contrastive learning strategy augmented by motion insights, integrating both global context and dynamic movement for a holistic model.
CLIP-FSAR\cite{wang2023clip} harnesses the robust generalization of CLIP \cite{Clip-origin2021}, employing encoders to process both text and images, subsequently benefiting from Transformer for comparison.

\section{Methodology}
\subsection{Problem Formulation}
In the normal few-shot action recognition setting, we define a full labeled dataset as $\mathcal{D}_S = \{x_i, y_i \}^{N_S}_{i=1} \in  \mathcal{X}_S \times \mathcal{Y}_S$, and a few labeled dataset as $\mathcal{D}_T = \{x_i, y_i \}^{N_T}_{i=1} \in  \mathcal{X}_T \times \mathcal{Y}_T$. FSAR aims to train a model 
in $\mathcal{D}_S$ and generate it to $\mathcal{D}_T$. Usually, meta-training and meta-testing are used for the FSAR.
In the FSAR, $\mathcal{Y}_S \perp \mathcal{Y}_T$. And the feature spaces are same, $\mathcal{X}_S = \mathcal{X}_T$.
In the field of cross-domain few-shot action recognition, the source dataset and the target dataset belong to different domains, just as $\mathcal{X}_S \neq \mathcal{X}_T$, and there is a significant domain gap between the different domains. Labels of different domains are also orthogonal.
The dataset $\mathcal{D}_S$ has enough labels and is used in the training stage.
We select some of the unlabeled data in the $\mathcal{D}_T$ as $\mathcal{D}_{u_{train}}$, $u_{train}$ means ``unlabeled target domain data for training''.
For the training, in each episode, $N$ classes with $K$ samples in $\mathcal{D}_S$ are sampled as ``support set'', and the samples from the rest of the videos in each category are sampled as ``query set'', just as $P$ samples are selected from each category to construct the ``query set''. Also, $N^{'}$ unlabeled samples in the $\mathcal{D}_{u_{train}}$ are selected for training.
For evaluation, 
We define a few labeled data in the $\mathcal{D}_T$ as $\mathcal{D}_{lt}$, $lt$ means ``labeled target domain data'', and the other data in the $\mathcal{D}_T$, are marked as $\mathcal{D}_{ut}$, $ut$ means ``unlabeled target data for testing''.
In each episode, $N$ classes with $K$ labeled samples in the $\mathcal{D}_{lt}$ are defined as ``support set'', and the data in the $\mathcal{D}_{ut}$ are called ``query set''. 
We can get $\mathcal{D}_{T} = \mathcal{D}_{u_{train}} \cup \mathcal{D}_{lt} \cup \mathcal{D}_{ut}$.
The goal of cross-domain few-shot action recognition is to train a model using $\mathcal{D}_S$ and $\mathcal{D}_{u_{train}}$ and generalize the model to $\mathcal{D}_T$.
In each task, we define the categories of ``support set'' as $\mathcal{C} = \{c_1, ..., c_N \}$, and we aim to
 classify a query video $q$ into $c_i \in \mathcal{C}$. 

\subsection{Overall framework}

\begin{figure*}
\centering
\includegraphics[width=17cm]{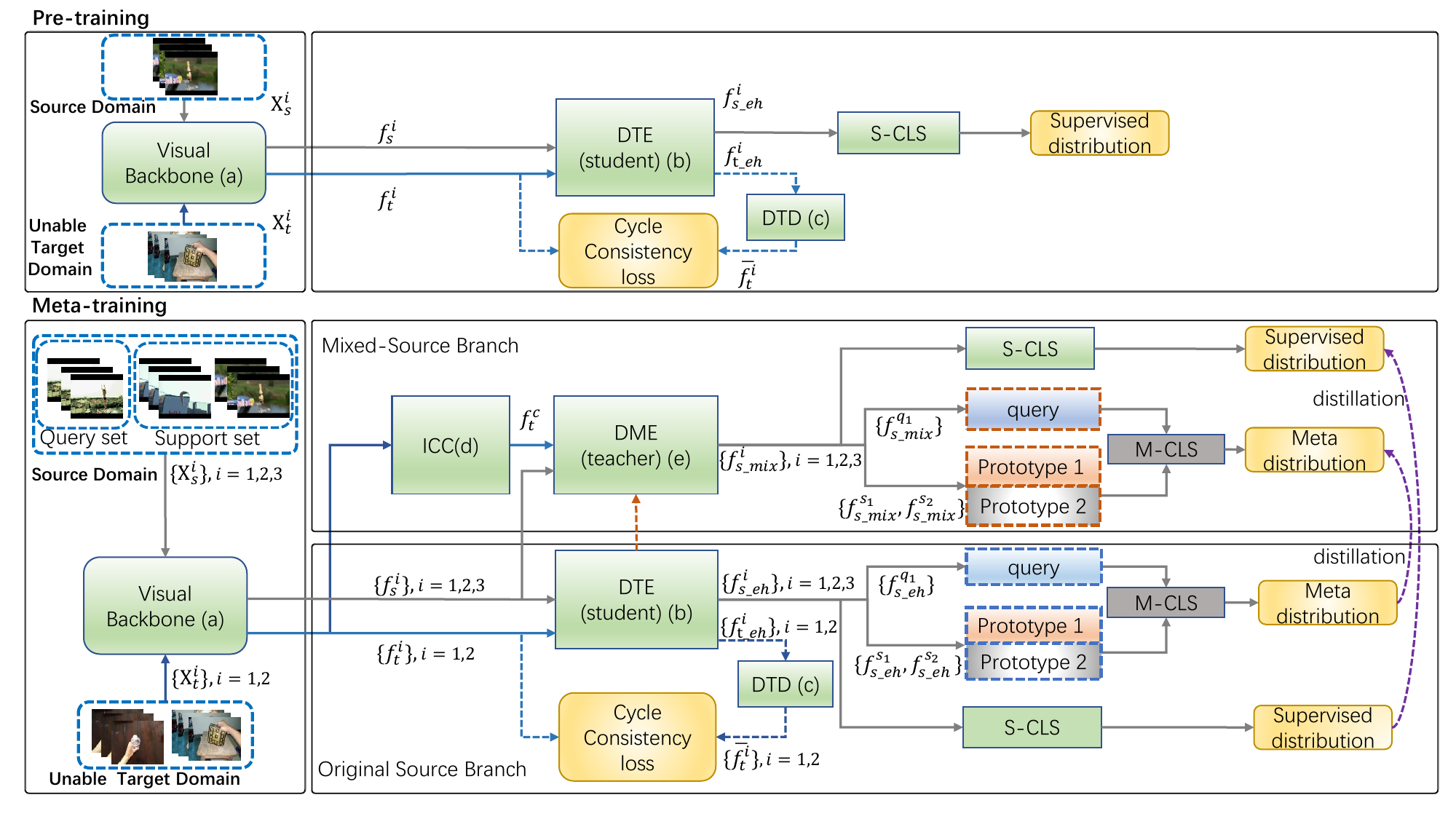}
\caption{The Framework of DMSD. In this illustration, the top part is the pre-training stage, and the bottom is the meta-training stage.
\textbf{(a)} Backbone of ResNet18 ($\Psi_{\phi}$).
\textbf{(b)} Domain Temporal Encoder. 
\textbf{(c)} Domain Temporal Decoder. 
\textbf{(d)} Instance Center Calibration. 
\textbf{(e)} Domain Mixer Encoder.
S-CLS means supervised classifier, and M-CLS means the meta-learning classifier.
(\textbf{Note}: 
The blue arrows represent unlabeled target features, while the brown arrows represent source features or mixed-source features. 
For simplicity, we use a 2-way 1-shot 1-query setting in the illustration. We use linear classifier as S-CLS and OTAM as M-CLS)
} 
\label{fig:2}
\end{figure*}

We present the framework for cross-domain few-shot action recognition, illustrated in Figure \ref{fig:2}. Our approach adopts a two-stage training strategy. 
In both of the stages, we employ ResNet18\cite{he2016deep-resnet} as the backbone $\Psi_{\phi}$ for the spatial feature extraction. To prevent potential biases from ImageNet pre-training parameters affecting cross-domain performance, we abstain from utilizing them for model initialization.
In the pre-training stage (refer to upper part of \Cref{fig:2}), we focus on creating a robust feature extractor for spatial and temporal dimensions using the same modules as the original-source branch in the upcoming meta-training stage. 
We rely on supervised learning and feature reconstruction in this stage.
In the meta-training stage (refer to the lower part of \Cref{fig:2}), following the backbone, we introduce an original-source branch that uses a Domain Temporal Encoder based on Transformer \cite{vaswani2017attention} for temporal feature extraction, feeding the spatiotemporal features to a supervised classifier. Also, a Domain Temporal Decoder and Cycle Consistency Loss are used for feature reconstruction, aiming for shared representations. Simultaneously, we integrate a meta-learning classifier into the original-source branch alongside the supervised one to enable swift adaptation to training samples via meta-learning.
Additionally, we incorporate a mixed-source branch that utilizes a Domain Mixer Encoder to shift the source domain data grounded on the unlabeled target domain in the temporal dimension, yielding mixed-source domain data. This branch also employs both supervised and meta-learning classifiers.
Finally, we employ dual distillation to align the classification probabilities of the original-source branch with those of the mixed-source branch, which contains target-domain information. We also update the Domain Mixer Encoder as the Exponential Moving Average (EMA) of the Domain Temporal Encoder.

Following the rule of TSN\cite{wang2016temporal-tsn}, frames are sampled from each video.
Assume that $X_{t}^i$ and $X_{s}^i$ come from the target domain and source domain, 
through the backbone $\Psi_{\phi}$, we denote the representation $\Psi_{\phi}(X_t^i)$ and $\Psi_{\phi}(X_s^i)$, where dimension is $M\times D \times H \times W$, $M$ is the number of frames, $D$ is the dimension of channels, $H$ and $W$ are the dimension of 2D features.
Then, we adopt average pooling on the spatial dimension $H \times W$ and get the representation $f_s^i$ and $f_t^i$, where the dimension is $M\times D$.

\subsection{Original-Source Branch}
The original-source branch is optimized via multitask learning \cite{caruana1997multitask}, concurrently mastering supervised classification, meta-learning classification, and unsupervised target feature reconstruction.
The aim is for the shared representation encoded by the branch to capture commonalities among these tasks, thereby extracting valuable information for cross-domain few-shot action recognition.

\subsubsection{Domain Temporal Encoder (DTE)} \label{section-DTE}
\begin{figure}[htbp]
    \centering
    \includegraphics[width=7cm]{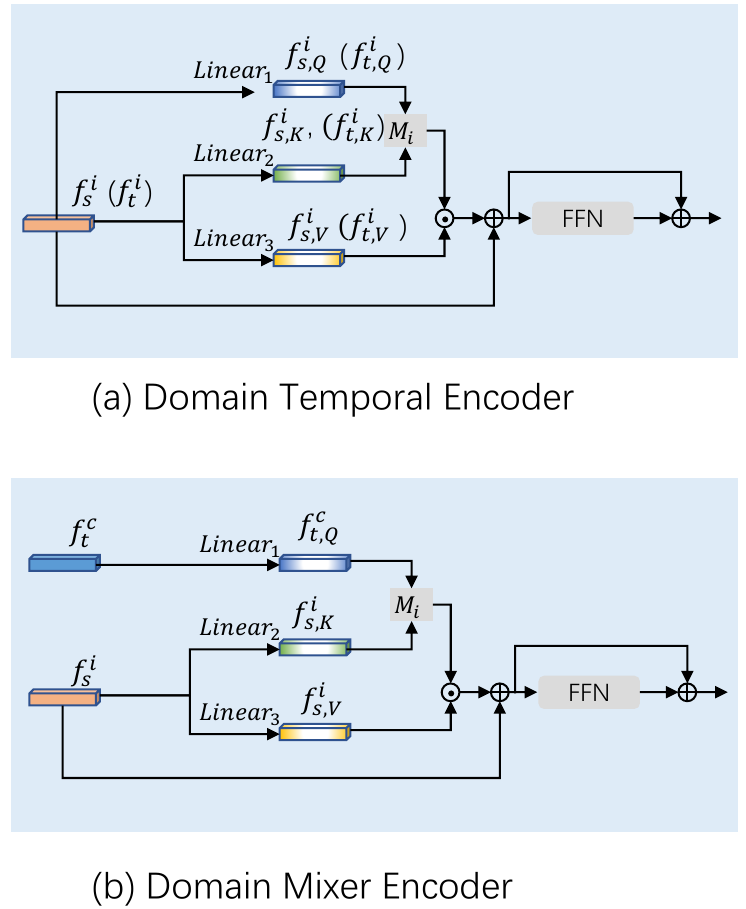}
    \caption{
(a) The structure of the Domain Temporal Encoder
(b) The structure of the Domain Mixer Encoder}
 \label{fig:3}
\end{figure}

Unlike images, which primarily contain spatial information, videos encapsulate a wealth of temporal dynamics. Capturing these temporal dependencies is pivotal for comprehensive video comprehension. Furthermore, findings from SEEN \cite{wang2023cross-crossDomainAction} highlight the significance of temporal relationships within videos for enhancing transfer and generalization capabilities in cross-domain few-shot action recognition.
We employ a Temporal Encoder based on the Transformer as a temporal modeling operation for the feature sequence. It enables the exploration of temporal structures and long-term dependencies within videos.
Referring to \Cref{fig:3}(a), we assume the video feature through the backbone is $f_x^i$ (\textbf{Note: $x$ is $s$ or $t$, meaning the video is from the source domain or target domain}).
Through 3 Linear operations, we obtain $f^i_{x,Q}$, $f^i_{x,K}$, and $f^i_{x,V}$ as the $Query$, $Key$ and $Value$.
Then, we utilize element-wise multiplication between $f^i_{x,Q}$ and $f^i_{x,K}$ to obtain the weights $M_i$. Subsequently, we reweigh $f_{x,V}^i$ by $M_i$. This process can be expressed as follows:
\begin{equation} \label{eq1}
       M_i = f^i_{x,Q} \odot  f^i_{x,K};  \quad  \tilde{f}^i_{x,V} =  f^i_{x,V} \odot M_i
\end{equation}
where $\odot$ denotes element-wise multiplication.
Let 
$\tilde{f}^i_{x,V}$ be the reweighted value features. To preserve the original visual information of the video, we utilize residual connections on the reweighted value features, defined as:

\begin{equation}  \label{eq2}
      f^{'i}_{x} =  \tilde{f}^i_{x,V} \oplus f^i_{x}
\end{equation}
where $\oplus$ denotes element-wise addition.
Through FeedForward and residual connections, we could get the final result.
\begin{equation}  \label{eq3}
      f^{i}_{x\_eh} =  f^{'i}_{x} \oplus FFN(f^{'i}_{x})
\end{equation}
Through this Temporal Encoder, features of the source domain are fed into two types of classifiers: supervised classifier and meta-learning classifier in each episode.
In meta-learning, for an N-way, K-shot setting with P-queries in each episode, the source domain feature set
$\{f_{s\_eh}^i\}, i \in \{1,2,..., N\times K + P \}$ 
should be divided into support set 
$\{f_{s\_eh}^{s_i}\}, i \in \{1,2,..., N \times K \}$ and query set 
$\{f_{s\_eh}^{q_i}\}, i \in \{1,2,..., P \}$.
Then, we can get the prototype $\{f_{s\_eh}^{s^c}\}, c \in \{c_1,c_2,..., c_N\}$ for each category of the support set according to the mean operation.
Given the source domain feature as $f_{s\_eh}^i$, 
we define $\mathcal{P}^s_i$ as the probability distribution obtained through the linear classifier (supervised classifier) with the dimension of output equal to the number of categories in the source domain dataset. 
\begin{equation}  \label{eqa}
\mathcal{P}^s_i = Linear(f_{s\_eh}^i)
\end{equation}
if the $f_{s\_eh}^i$ only belongs to the query set of the source domain dataset,
we define $\mathcal{P}^m_i$ as the probability distribution obtained through the OTAM \cite{cao2020few-otam} (meta-learning classifier) in each episode.
\begin{equation}  \label{eqab}
\mathcal{P}^m_i = OTAM(f_{s\_eh}^i, \{f_{s\_eh}^{s^c}\}), c \in \{c_1,c_2,..., c_N\}
\end{equation}
Using the  $\mathcal{P}^s_i$ and the label of $X^i_s$, through the Cross-Entropy, we can get the supervised loss $L^i_{super}$.
\begin{equation}  \label{eqac}
L^i_{super} = -\sum_{w\in W } \mathbb{1} (Y_i=w)  \cdot \log(\mathcal{P}^s_i[w])
\end{equation}
where $W$ is the set of the whole category in the source domain, $Y_i$ is the label of $X^i_s$. Using the  $\mathcal{P}^m_i$ and label of $X^i_s$, through the Cross-Entropy, we can get the meta loss $L^i_{meta}$.
\begin{equation}  \label{eqad}
L^i_{meta} = -\sum_{c\in C }  \mathbb{1} (Y_i=c) \cdot \log(\mathcal{P}^m_i[c])
\end{equation}
where $C = \{c_1,c_2,..., c_N\}$ per episode, $Y_i$ is the label of $X^i_s$.

Additionally, through this Domain Temporal Encoder, features of the target domain are fed into a Domain Temporal Decoder.

\subsubsection{Domain Temporal Decoder (DTD) and Cycle Consistency Loss}
In Figure \ref{fig:2}, the feature of target domain video $X^i_t$ processed by the Temporal Encoder is denoted as $f^i_{t\_eh}$. We then employ a Domain Temporal Decoder designed to decode features from the Domain Temporal Encoder, with the goal of reconstructing the spatial features of ResNet18. During training, Cycle Consistency Loss is computed to guide the model in accurately reconstructing the spatial features. This ensures minimal information loss during encoding and decoding, providing continuous self-supervision throughout training and yielding a robust temporal encoding for the target domain.

\begin{equation}  \label{eq4}
   \overline{f_t^i} = Decoder(f_{t\_eh}^{i})
\end{equation}
where the structure of the Decoder is similar to that of the Temporal Encoder.
Then, we use the Euclidean distance between the $\overline{f_t^i}$ and $f_t^i$ as the Cycle Consistency Loss.
\begin{equation}  \label{eq5}
   L_{con}^i = \frac{1}{D}|| \overline{f_t^i}-f_t^i||^2
\end{equation}
where $D$ is the dimension of the frame feature.

\subsection{Mixed-Source Branch}
In prior research \cite{samarasinghe2023cdfsl-crossDomainAction-cdfsl-v, islam2021dynamic-cdfs-dynamic-distillation}, the integration of target domain information involved computing the distillation loss between teacher and student networks. Specifically, pseudo labels were generated using the sharp teacher output for weakly augmented target images. These labels were then used in conjunction with the student output on a strongly augmented version of the same images to calculate the consistency distillation loss. Concurrently, the teacher network was updated as a moving average of the student's weights.

Our approach leverages target domain information to enrich the source domain by introducing the novel concept of a mixed-source domain. In each episode, we first calculate the weighted aggregation center of the unlabeled target domain to represent it effectively. We then fuse this weighted center with the features of the source domain through a cross-transformer in the temporal dimension, thus creating the mixed-source domain. Since the labels of the mixed-source domain align with those of the source domain, both supervised and meta-learning classifications can also be performed within the mixed-source branch.

\subsubsection{Instance Center Calibration (ICC)} \label{IR}
We aim to enhance the source domain in each episode using target domain features. We assume that a group of randomly sampled target domain feature set $\{f_t^i\}$, $i \in \{1,2,..., N^{'}\}$ per episode can reflect the distribution of the target domain. Typically, mean aggregation represents a group of samples, but this approach may introduce certain inductive biases. To address this issue, we propose calibrating the mean by considering the similarity between target instances. By incorporating similarity into the aggregation process, we aim to refine the representation of the target domain, ensuring it captures the nuanced characteristics more effectively. Refer to \Cref{eq6,eq7,eq8}.
\begin{equation}  \label{eq6}
   cos\_sim(i,j) = \frac{<f_t^i, f_t^j> }{||f_t^i|| \cdot ||f_t^j||}
\end{equation}

\begin{equation}  \label{eq7}
   \alpha_i = \frac{1}{N^{'}-1} \sum\limits_{j=1 }^{N^{'}-1} cos\_sim(i,j)
\end{equation}

\begin{equation}  \label{eq8}
   f^c_t = \frac{1}{N^{'}-1} \sum\limits_{i=1 }^{N^{'}-1} \alpha_i f_t^{i}
\end{equation}
where the $\alpha_i$ represents the adaptive weight for the $i$th instance, which is obtained by averaging the similarity scores between the $i$th instance and other instances within the target domain.
The $N^{'}$ is the number of target samples in each episode.
With these weights, we can get the calibrated instance center as $f_{t}^c$, ensuring a more refined representation that encapsulates the subtle nuances of the target domain.

\subsubsection{Domain Mixer Encoder (DME)}
The Domain Mixer Encoder shares a structural resemblance with the Domain Temporal Encoder from the original-source branch, but it employs cross-attention rather than self-attention for video processing. It refines the source domain samples by adopting the target domain center, calculated through Instance Center Calibration to create the $Query$ and utilizing the source domain samples to create $Key$ and $Value$. This alignment reduces distribution differences between the source and target domains, improving performance on the target domain. 
The Domain Mixer Encoder, leveraging cross-transformer, better captures the relationships between domains, enabling more effective generalization to unseen target domain data. As depicted in \Cref{fig:3}(b), its computational process is similar to that of the DTE. Refer to \Cref{eq1,eq2,eq3}. Through the DME, we can get the mixed-source feature set 
$ \{f_{s\_mix}^i\}, i \in \{1,2,..., N\times K + P \}$. 
For meta-learning, the mixed-source feature set should be divided into a support set 
$ \{f_{s\_mix}^{s_i}\}, i \in \{1,2,..., N\times K \}$ and a query set 
$\{f_{s\_mix}^{q_i}\}, i \in \{1,2,...,P \}$.
Then, we can get the prototype $\{f_{s\_mix}^{s^c}\}, c \in \{c_1,c_2,.., c_N\}$ for each category of the support according to the mean operation.
The definitions of $N$, $K$, and $P$ are the same as those in DTE.

For the mixed-source feature $f_{s\_mix}^i$, similar to \Cref{section-DTE}, we define $\mathcal{Q}^s_i$ as the probability distribution of supervised classification using a linear classifier and $\mathcal{Q}^m_i$ as the probability distribution of meta-learning classification using OTAM\cite{cao2020few-otam}.

\subsection{Dual Distillation for learning the target}
In the meta-training, we encourage extracting richer semantic information from the video's mixed-source representations. 
To achieve this goal, we impose consistency constraints on the classification between mixed-source features and original-source features. 
We define the Domain Mixer Encoder in the mixed-source branch as the teacher and the Domain Temporal Encoder in the original-source branch as the student.
Given a source domain video $X^i_s$, we utilize both the original-source and mixed-source branches to generate classification predictions. These predictions are of two types: $N$-way classification of each episode and supervised classification across all predicted categories.
To quantify the consistency between the two types of predictions, we employ KL divergence to approximate the similarity of the probability distribution from the original-source branch relative to that of the mixed-source branch, as shown in Equation \ref{eq9}:

\begin{equation} \label{eq9}
    \begin{split}
        L_{m}^i= \sum\limits_{j=1}^{N}  \mathcal{Q}^m_{i,j} \log \frac{\mathcal{Q}^m_{i,j} } { \mathcal{P}^m_{i,j} } \\
        L_{s}^i= \sum\limits_{j=1}^{\tilde{N}}  \mathcal{Q}^s_{i,j} \log \frac{\mathcal{Q}^s_{i,j} } { \mathcal{P}^s_{i,j} }
    \end{split}
\end{equation}
where the $\mathcal{P}^s_i$ and $\mathcal{P}^m_i$ represent the probability distributions of predictions from the original-source branch, the $\mathcal{Q}^s_i$ and $\mathcal{Q}^m_i$ represent the probability distributions of predictions from the mixed-source branch. $m$ means meta-learning, $s$ means supervised learning. $N$ is the number of categories in each episode, and $\tilde{N}$ is the number of the whole categories.

\subsection{Momentum Update}
In the proposed method, the Domain Mixer Encoder in the mixed-source branch mirrors the structure of the Domain Temporal Encoder in the original-source branch. 
A straightforward approach would be to update both branch networks simultaneously based on the loss function. However, this introduces additional learnable parameters and can lead to inefficiencies during training.
To address these challenges, we suggest learning domain knowledge during meta-training by updating the Domain Mixer Encoder in the mixed-source branch to align with the exponential moving average (EMA) of the Domain Temporal Encoder in the original-source branch. This method enables more effective learning of domain knowledge and further enhances model performance. Specifically, we update the parameters of the Mixer Encoder as follows:
\begin{equation} \label{eq10}
\theta_t \longleftarrow \alpha \theta_t + (1-\alpha) \theta_s
\end{equation}
where $\theta_t$ is the parameters of Domain Mixer Encoder, and  $\theta_s$ is the parameters of Domain Temporal Encoder.

\subsection{Training loss and prediction}
In the pre-training stage, the loss is defined as:
\begin{equation} \label{eq11}
 L_{total}^p =
             \alpha_1 \sum\limits_{i=1}^{N^{'}} L^i_{con}    +  
             \alpha_2 \sum\limits_{i=1}^{N \times K+P}   L^i_{super}
\end{equation}
In the meta-training stage, the total loss is defined as:
\begin{equation} \label{eq12}
\begin{split}
 L_{total}^m =
             &\alpha_1\sum\limits_{i=1}^{N^{'}} L^i_{con}    +  
             \alpha_2\sum\limits_{i=1}^{P}   L^i_{meta}   + \\
             &\alpha_3\sum\limits_{i=1}^{N \times K+P}   L^i_{super}  +
             \alpha_4\sum\limits_{i=1}^{P}   L^i_{m}  +
             \alpha_5\sum\limits_{i=1}^{N \times K+P}   L^i_{s}
\end{split}
\end{equation}
where $N^{'}$ is the number of unlabeled samples in the target domain in each episode. 
$N$ is the category number of each episode in the source domain, $K$ is the number of support samples per category, and $P$ is the number of query samples per episode. 
Additionally, $\alpha_i$, where $i=1,2,3,4,5$, represents the hyper-parameters used to balance the losses. 

For prediction, our goal is to classify the target domain dataset. Features extracted by the Domain Temporal Encoder are sent to OTAM \cite{cao2020few-otam} for comparison. Importantly, our proposed method does not entail fine-tuning.

\section{Experiments}
\subsection{Datasets} \label{sec: datasets}

In the realm of regular few-shot action recognition, datasets such as UCF101 \cite{soomro2012ucf101}, HMDB51 \cite{kuehne2011hmdb}, Kinetics \cite{carreira2017quo-kinetics}, and SSV2 \cite{goyal2017something} are utilized for model training and evaluation.
In SEEN \cite{wang2023cross-crossDomainAction}, experimental setups use both Kinetics-Small and Kinetics-Full as the source datasets. Both datasets comprise 64 classes for training, with the distinction being that Kinetics-Small contains 100 samples per class, while Kinetics-Full encompasses all annotated videos from Kinetics for each class, totaling approximately 52K samples in its training set.
In the CDFSL-V experiments \cite{samarasinghe2023cdfsl-crossDomainAction-cdfsl-v}, Kinetics-100 and Kinetics-400 are employed. Kinetics-100 resembles Kinetics-Small in SEEN.

\begin{figure}
\centering
\includegraphics[width=8cm]{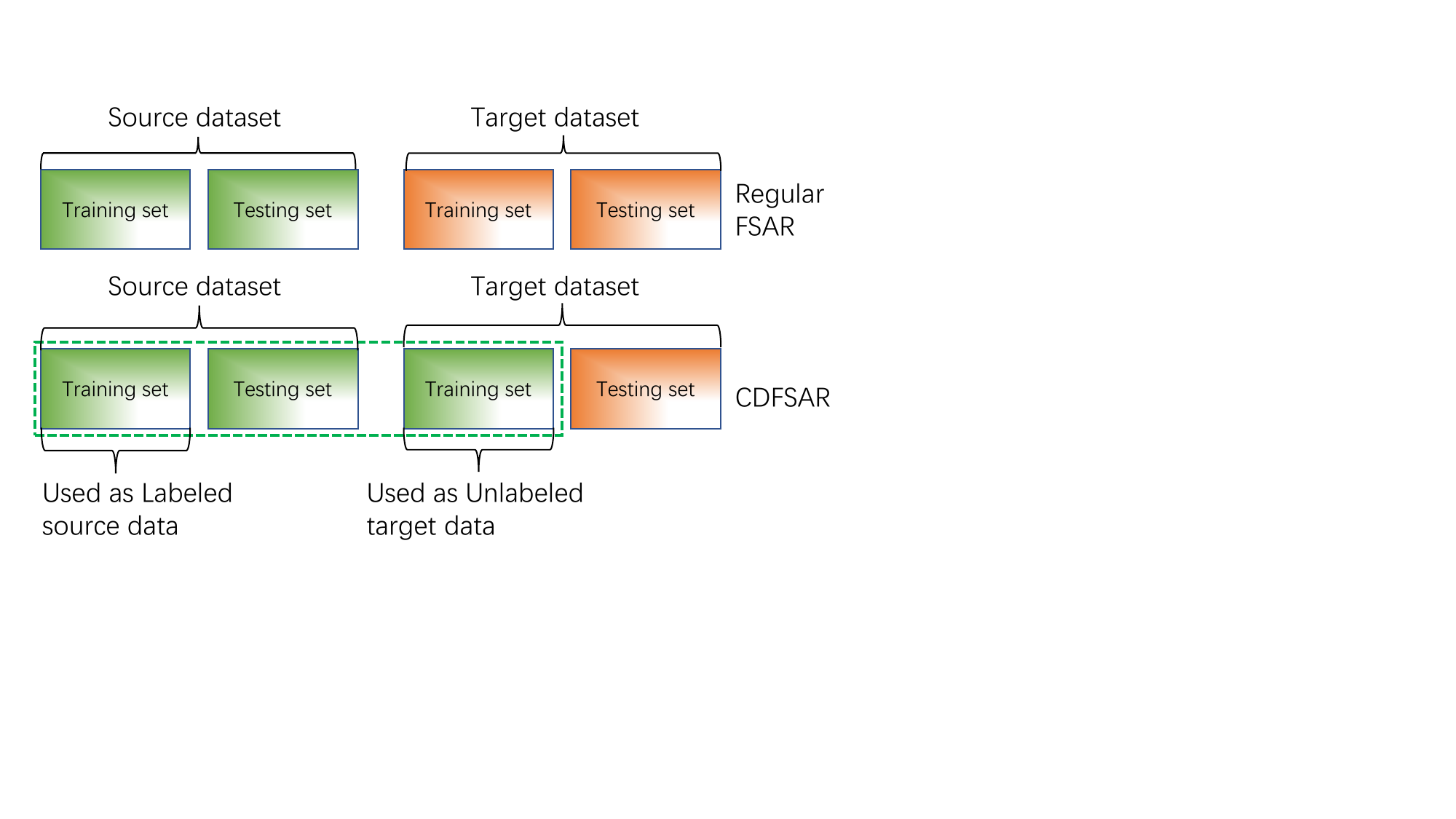}
\caption{\label{fig:datasetSplit} The rule for dataset split for cross-domain Few-shot Action Recognition is based on the regular FSAR split method. 
} 
\label{fig:4}
\end{figure}

\textbf{Source and the Target Domain setting}:

\textbf{Main setting.}
In most experiments, we use the Kinetics as the source domain.
To ensure an effective comparison with previous works, we adopt the Kinetics-Small of SEEN in the methods comparison experiment, defined as Kinetics100-Small.
In ablation experiments, we simultaneously utilize Kinetics-Full from SEEN, defined as Kinetics100-Full. 
Additionally, we utilize Kinetics200/300/400-Small, wherein each class maintains 100 samples.
Our target domain datasets encompass UCF101, HMDB51, SSV2, Diving48 \cite{li2018resound}, and RareAct \cite{miech2020rareact}. 
For FSAR:
UCF101 consists of 101 action classes, following the split in ARN \cite{zhang2020few-ARN}. It comprises 70 classes for training, 10 for validation, and 21 for testing, with $9154/1421/2745$ videos for $train/val/test$, respectively.
HMDB51 adheres to the split rule of ARN \cite{zhang2020few-ARN} for FSAR, featuring 31 training classes, 10 verification classes, and 10 testing classes, with $4280/1194/1292$ videos for $train/val/test$.
SSV2 utilizes the OTAM \cite{cao2020few-otam} split method for FSAR, similar to the SSV2-CMN split method. It comprises 64 classes for training, 12 for validation, and 24 for testing, with all samples within each class used. The OTAM split method contains $77500/1926/2854$ videos for $train/val/test$. 
For UCF101, HMDB51, and SSV2, we repurpose the meta-training set as unlabeled target data, which will be employed during training alongside the source domain, see \Cref{fig:4}.
For Diving48 and RareAct, we adhere to the split provided by CDFSL-V.

\textbf{Other setting.} In one of the ablation studies, we sequentially use each dataset as the source domain while treating the other datasets as target domains to investigate the related patterns in cross-domain problems.

\subsection{Implementation Details}
Following TSN \cite{wang2016temporal-tsn} and previous methods, we sparsely and uniformly sample several ${(i.e., T=8)}$ frames for each video. We then horizontally flip each frame and randomly crop the center region of 224 × 224 for training data augmentation. We use ResNet18 \cite{he2016deep-resnet} as the backbone, but without pre-trained parameters on ImageNet \cite{deng2009imagenet}. Additionally, we remove the final fully connected layer to obtain frame-level features, which have 512 dimensions.
During the pre-training, we sample 10,000 episodes. In the meta-training stage, we sample from 5,000 to 15,000 episodes. In the meta-testing, we sample 10,000 episodes and calculate the average accuracy of each episode.
We use the Adam optimizer with a learning rate of 0.001 and train our model on the Nvidia 3090 GPU. During training, we average gradients and perform back-propagation once every 16 iterations.
For testing, we use only the center crop to augment the video. Fine-tuning with a few labeled samples of the target domain is unnecessary, and meta-comparison suffices for testing.

\subsection{Comparison with previous works}
Our work mainly compares with the current SEEN \cite{wang2023cross-crossDomainAction} and CDFSL-V \cite{samarasinghe2023cdfsl-crossDomainAction-cdfsl-v} methods, and we also compare with several CDFSL  transfer-based methods. Also, we introduce several FSAR methods for comparison, such as STRM \cite{thatipelli2022spatio-strm}, TRX \cite{perrett2021temporal-trx}, and HYRSM \cite{wang2022-hybrid}. 
For each target domain, we report the accuracy of our model under the settings of 1-shot and 5-shot. 
It can be seen that our model outperforms the current cross-domain few-shot action recognition methods on all datasets.
Compared to SEEN, under the 5-shot setting:
The accuracy of UCF101 has increased from $79.6\%$ to $81.90\%$. 
The accuracy of HMDB51 has increased from $52.8\%$ to $54.90\%$. 
The accuracy of SSV2 has increased from $31.2\%$ to $32.10\%$. 
The accuracy of Diving48 has increased from $40.9\%$ to $42.28\%$. 
The accuracy of RareAct has increased from $50.2\%$ to $53.30\%$.
Under 1-shot setting:
The accuracy of UCF101 has increased from $64.6\%$ to $65.0\%$. 
The accuracy of HMDB51 has increased slightly from $39.5\%$ to $39.62\%$. 
The accuracy of SSV2 has increased from $26.2\%$ to $26.78\%$. 
The accuracy of Diving48 has increased from $31.23\%$ to $34.20\%$. 
The accuracy of RareAct has increased from $34.28\%$ to $37.50\%$.

We present a baseline for the DMSD, consisting only of the ResNet18 and the OTAM comparison method. The baseline approach is rooted in the classic meta-learning method. While it generally shows lower accuracy than most transfer methods STARTUP and Dynamic Distillation, it outperforms models like TRX, STRM, and HYSRM, designed for same-domain few-shot action recognition. Compared to our baseline, DMSD demonstrates a significant improvement in accuracy and outperforms the transfer methods.

From the results summarized in the \Cref{tab:1}, we can get:
(1) Regular meta-learning methods for FSAR based on metrics perform inferiorly compared to transfer-based methods.
(2) Our model outperforms both regular meta-learning methods and transfer-based methods.
(3) The accuracies of CDFSL-V under our implementation are similar to the published results \cite{samarasinghe2023cdfsl-crossDomainAction-cdfsl-v} but consistently lower than other methods, potentially due to its use of VideoMAE as the backbone.

\begin{table*}[htbp]
  \centering
  \caption{
5-way 5-shot and 1-shot Cross-Domain FSAR accuracy using Kinetics100-Small as the source domain dataset. The best results are highlighted in bold. ``-'' indicates unavailable results, and $\diamond$ denotes our implementation. NOTE: The partial comparison results in the experiment are sourced from Table-1 in SEEN \cite{wang2023cross-crossDomainAction}. Pretrained parameters are not used in the backbone of each method.
  }
  \scalebox{0.92}{
    \begin{tabular}{l|l|cc|cc|cc|cc|cc}
    \toprule
    \multicolumn{1}{c|}{\multirow{2}[2]{*}{Method}} & \multicolumn{1}{c|}{\multirow{2}[2]{*}{Reference}} & \multicolumn{2}{c|}{UCF101} & \multicolumn{2}{c|}{HMDB51} & \multicolumn{2}{c|}{SSV2} & \multicolumn{2}{c|}{Diving48} & \multicolumn{2}{c}{RareAct} \\
          &       & \multicolumn{1}{l}{1-shot} & \multicolumn{1}{l|}{5-shot}  & \multicolumn{1}{l}{1-shot} & \multicolumn{1}{l|}{5-shot} & \multicolumn{1}{l}{1-shot} & \multicolumn{1}{l|}{5-shot} & \multicolumn{1}{l}{1-shot} & \multicolumn{1}{l|}{5-shot} & \multicolumn{1}{l}{1-shot} & \multicolumn{1}{l}{5-shot} \\
    \midrule
    STARTUP \cite{phoo2020self-cdfs-startup} & ICLR'2021   & 64.1 & 79.4 & 38.3 & 50.7&25.2 & 29.8& --  & -- &  --  & --\\
    Dynamic Distillation \cite{islam2021dynamic-cdfs-dynamic-distillation}&NIPS'2021 & 63.9 & 77.0&  37.5& 49.9& 24.5 & 28.7&  --     & --&  --  & -- \\
    TRX \cite{perrett2021temporal-trx}  & CVPR’2021  &44.5& 67.3  & 28.2 &45.1 & 22.8&27.7 & 31.17$^{\diamond}$ & 40.84$^{\diamond}$ & 32.2$^{\diamond}$&50.21$^{\diamond}$\\
    STRM \cite{thatipelli2022spatio-strm} & CVPR’2022  &45.0& 70.6     & 27.5 &45.6 & 23.1& 28.9 & 31.72$^{\diamond}$  & 41.92$^{\diamond}$ & 31.58$^{\diamond}$      &49.27$^{\diamond}$\\
    HYSRM \cite{wang2022-hybrid} & CVPR’2022  & 56.18$^{\diamond}$ &69.96$^{\diamond}$& 32.12$^{\diamond}$  &48.9$^{\diamond}$ & 24.6$^{\diamond}$ & 28.9$^{\diamond}$ & 30.16$^{\diamond}$ & 38.86$^{\diamond}$  &32.16$^{\diamond}$ &  50.72$^{\diamond}$ \\
   CDFSL-V \cite{samarasinghe2023cdfsl-crossDomainAction-cdfsl-v} & ICCV'2023 &38.29$^{\diamond}$  &45.17$^{\diamond}$ &  20.93$^{\diamond}$ &30.11$^{\diamond}$ & 16.01$^{\diamond}$ & 17.92$^{\diamond}$ & 17.21$^{\diamond}$ &23.77$^{\diamond}$ & 23.56$^{\diamond}$ &34.94$^{\diamond}$ \\
    SEEN \cite{wang2023cross-crossDomainAction} & CVIU'2023 & 64.6 &79.6 &39.5& 52.8 & 26.2 &31.2   &$31.23^{\diamond}$&$40.9^{\diamond}$ &$34.28^{\diamond}$ &$50.2^{\diamond}$  \\
    \midrule    \rowcolor{gray!15}  Baseline  & -- & 47.89 &76.86&  32.76  & 50.76   &22.28 &  28.85 &30.26 & 39.24&29.98 & 50.20   \\
    \rowcolor{gray!15}  DTMV(Ours)  & -- & \textbf{65.0} &\textbf{81.90}&  \textbf{39.62}  & \textbf{54.90}    &\textbf{26.78} &  \textbf{32.10} &\textbf{34.20} & \textbf{42.28}&\textbf{37.50}  &   \textbf{53.30}     \\
    \bottomrule
    \end{tabular}%
    }
  \label{tab:1}%
\end{table*}%

\subsection{Ablation Study}

\subsubsection{Impact of source domain settings}

\begin{table*}[htbp]
  \centering
  \caption{comparison of the results under different source domain settings. The details of the settings can be referred to in \Cref{sec: datasets}.
  }
    \begin{tabular}{l|ll|ll|ll|ll|ll}
    \toprule
    \multicolumn{1}{c|}{\multirow{2}[2]{*}{Soure Domain}} & \multicolumn{2}{c|}{UCF101} & \multicolumn{2}{c|}{HMDB51} & \multicolumn{2}{c|}{SSV2} & \multicolumn{2}{c|}{Diving48} & \multicolumn{2}{c}{RareAct} \\
          & 1-shot & 5-shot & 1-shot & 5-shot & 1-shot & 5-shot & 1-shot & 5-shot & 1-shot & 5-shot \\
    \midrule
    Kinetics100-Small &65.0 & 81.90 &36.92 & 54.90 & 26.78 & 32.10 & 34.20 & 42.28 & 37.50 &53.30\\
    Kinetics100-Full  &76.90 & 84.12 &37.22 & 56.25& 27.80& 35.30 & 33.91 & 45.42 & 38.90  & 54.41\\
    \bottomrule
    \end{tabular}%
  \label{tab:2}%
\end{table*}%

\begin{figure}[htbp]
    \centering
    \includegraphics[width=8cm]{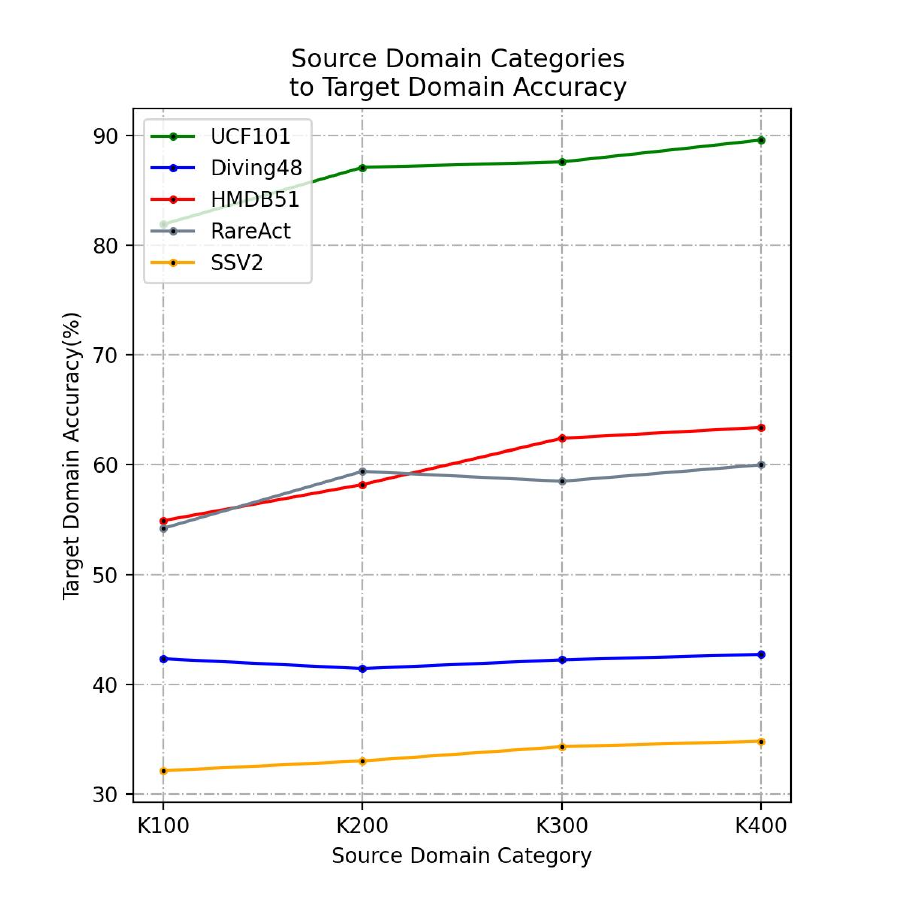}
    \caption{Accuracies of five target domain datasets with varying category sizes of the source domain.
    }
\label{fig:5}
\end{figure}

As illustrated in \Cref{tab:2}, a larger source dataset proves beneficial for acquiring more robust representations and enhancing generalization abilities. Increased samples per class are correlated with higher accuracy rates. We can see that the accuracy under Kinetics100-Full is higher than that under Kinetics100-Small. 
For experimentation, we selected 100, 200, 300, and 400 categories from the Kinetics, and each category contains 100 samples as the source domains. Results depicted in \Cref{fig:5} illustrate that as the category number in the source domain increases, the recognition accuracy in the target domain consistently rises. It underscores the positive correlation between the category number of the source domain and the model's generalization ability.
We can see the improvement in Diving48 is modest, increasing by less than $1\%$ when expanding from 100 to 400 categories in the Kinetics dataset. It can be attributed to the fine-grained nature of the Diving48 database, where the distinctions between different action categories are minimal. Despite the greater diversity in the source domain data, which is the diversity of a coarser grain, the transferability to Diving48 does not significantly improve.

\subsubsection{Analysis of Model Components}

In this subsection, we design and conduct comparative experiments on SSV2 and Diving48 to evaluate the performance of different branch selection settings, classification mode settings, and distillation mode settings for our model, aiming to demonstrate the effectiveness of our model architecture. To ensure clarity in the analysis, we maintain the use of Cyclic Consistency Loss in these ablation experiments.
We evaluate performance using both 5-shot and 1-shot settings, as shown in \Cref{tab:3}:
(1) When using only the original-source branch, combining both supervised loss and meta-loss yields higher accuracy than using only one type of loss.
(2) The accuracy of models using only the original-source branch is lower than when incorporating the mixed-source branch and utilizing it as the teacher network during training.
(3) When utilizing both branches, combining supervised loss and meta-loss results in higher few-shot classification accuracy compared to using only one type of loss.
(4) Using both supervised loss and meta-loss, models employing dual distillation achieve higher accuracy than those using only supervised results or only meta-learning results for distillation.
Combining the findings from \Cref{tab:3} and the above summaries, we can conclude that employing both supervised loss and meta-learning loss, along with dual distillation from the mixed-source branch to the original-source branch, can yield better results for the model.

\begin{table*}[htbp]
  \centering
  \caption{The ablation study about the influence of each component. \ding{51} indicates that the component is used, and vice versa.
  This experiment is for meta-training. In the pre-training and meta-training stages, we keep using Cycle Consistency Loss.
  }
    \begin{tabular}{c|cc|cc|cc|cc|cc}
    \toprule
          \multicolumn{1}{c|}{\multirow{2}[2]{*}{No.}} & \multicolumn{2}{c|}{Branch select} & \multicolumn{2}{c|}{Classify loss} & \multicolumn{2}{c|}{Distillation} & \multicolumn{2}{c|}{SSV2} & \multicolumn{2}{c}{Diving48} \\
          & \multicolumn{1}{l}{Original} & \multicolumn{1}{l|}{Mixed} & \multicolumn{1}{l}{Supervised } & \multicolumn{1}{l|}{Meta} & \multicolumn{1}{l}{Supervised } & \multicolumn{1}{l|}{Meta} & \multicolumn{1}{l}{1-shot} & \multicolumn{1}{l|}{5-shot} & \multicolumn{1}{l}{1-shot} & \multicolumn{1}{l}{5-shot} \\
    \midrule
     (1)     & \ding{51}      & \ding{55}      &\ding{51}       &\ding{55}       & --  & --      &  23.78 &  29.92 & 31.76 & 41.24 \\
     (2)     & \ding{51}      & \ding{55}      &\ding{55}       &\ding{51}       & --  & --      &  24.12 &  30.16  & 32.18  & 41.76  \\
     (3)     & \ding{51}      & \ding{55}      &\ding{51}       &\ding{51}       & --  & --      &  23.9 &  31.28  & 32.74  & 41.52 \\
    \midrule
     (4)     & \ding{51}      & \ding{51}      & \ding{51}      & \ding{55}       & \ding{51}      & \ding{55} & 24.18 &30.46 &32.04 & 41.02 \\
     (5)     & \ding{51}      & \ding{51}      & \ding{55}      & \ding{51}      & \ding{55}      & \ding{51}  &24.70  &30.44& 32.32 &  41.24 \\
     (6)     & \ding{51}      & \ding{51}      & \ding{51}      & \ding{51}        & \ding{51}      & \ding{55}  &25.20&31.28 & 33.02  & 41.40   \\
     (7)     & \ding{51}      & \ding{51}      & \ding{51}      & \ding{51}        & \ding{55}      & \ding{51}  & 24.82  &31.40 & 33.26      & 41.88   \\
     (8)     & \ding{51}      & \ding{51}      & \ding{51}      & \ding{51}        & \ding{51}      & \ding{51}  &   26.78    & 32.10 &  34.20     &  42.28 \\

    \bottomrule
    \end{tabular}%
  \label{tab:3}%
\end{table*}%

\subsubsection{The importance of Cycle Consistency Loss}
As shown in \Cref{tab:4}, omitting the Cycle Consistency Loss loss leads to the lowest accuracy for our model. Interestingly, models that incorporate Cycle Consistency Loss only during meta-training achieve slightly higher accuracies than those that introduce it solely during pre-training. Notably, the model that integrates Cycle Consistency Loss throughout the training process achieves the highest accuracy. This observation suggests that effective feature extraction, which aligns both target and source domains, requires comprehensive training of both the Encoder and Decoder.

\begin{table*}[htbp]
  \centering
  \caption{The ablation study is designed to evaluate the influence of Cycle Consistency Loss empirically. Utilizing the Kinetics100-Small dataset as the source domain, assessments are conducted on the RareAct and HMDB51 datasets as the target domains. Both pre-training and meta-training strategies are retained, and the dual distillation mechanism is integrated.
  }
    \begin{tabular}{c|c|c|cc|cc}
    \toprule
    \multicolumn{1}{c|}{\multirow{2}[2]{*}{NO.}} & \multicolumn{1}{c|}{\multirow{2}[2]{*}{\makecell{Cycle Consistency Loss\\ in Pre-training}}} & \multicolumn{1}{c|}{\multirow{2}[2]{*}{\makecell{Cycle Consistency Loss\\ in Meta-training}}} & \multicolumn{2}{c|}{RareAct} & \multicolumn{2}{c}{HMDB51} \\
          &       &       & \multicolumn{1}{l}{1-shot} & \multicolumn{1}{l|}{5-shot} & \multicolumn{1}{l}{1-shot} & \multicolumn{1}{l}{5-shot} \\
    \midrule
       (1)   &\ding{55}      & \ding{55}      & 35.82 &52.84   & 37.90 & 52.30\\
       (2)   &\ding{55}      & \ding{51}      & 36.40& 52.46 & 38.76  &  52.54   \\
       (3)   &\ding{51}      & \ding{55}      & 36.34 & 53.12 &38.14& 52.30 \\
       (4)   &\ding{51}      & \ding{51}      &37.50 & 53.30  & 39.62  & 54.90    \\
    \bottomrule
    \end{tabular}
  \label{tab:4}
\end{table*}

\subsubsection{The effect of Instance Center Calibration (ICC)}

To validate the efficacy of Instance Center Calibration, we conducted experiments on the Diving48, RareAct, and SSV2 datasets. We juxtapose the utilization of the calibration center against the mean operation as the representation method for the target domain. Likewise, experiments are executed under both 1-shot and 5-shot settings. As illustrated in Table \Cref{tab:icc}, overall, employing the calibration center yields superior outcomes in both the 5-shot and 1-shot scenarios. Here, \ding{51} denotes the presence of the ICC, while \ding{55} signifies the utilization of mean operation instead of ICC.

\begin{table}[htbp]
  \centering
  \caption{The ablation study to verify the Instance Center Calibration (ICC), the source domain dataset is Kinetics100-Small.}
    \begin{tabular}{l|rr|rr|rr}
    \toprule
    \multicolumn{1}{c|}{\multirow{2}[1]{*}{ICC}} & \multicolumn{2}{c|}{Diving48} & \multicolumn{2}{c|}{RareAct} & \multicolumn{2}{c}{SSV2} \\
          & \multicolumn{1}{l}{1-shot} & \multicolumn{1}{l|}{5-shot} & \multicolumn{1}{l}{1-shot} & \multicolumn{1}{l|}{5-shot} & \multicolumn{1}{l}{1-shot} & \multicolumn{1}{l}{5-shot} \\
    \midrule
    \ding{55}    &33.91 & 42.0   &36.80 & 52.44 & 25.26   &31.62  \\
    \ding{51}   &34.20 & 42.28 & 37.50 & 53.30 & 26.78 & 32.10  \\
    \bottomrule
    \end{tabular}%
  \label{tab:icc}%
\end{table}%

\subsection{Different Source Domain Testing}

We use each dataset as the source domain, with the remaining datasets serving as the target domains. As shown in \Cref{table:sandt}, except for UCF101, in each column, whether under the 1-shot or 5-shot setting, the recognition performance is optimal when the source domain dataset matches the target domain. 
Overall, using Kinetics as the source domain yields the best transfer performance to other datasets, while using Diving48 as the source domain yields the worst. The reason is that Kinetics has greater diversity in its video data, whereas Diving48 consists solely of diving actions. Although the SSV2 dataset is also quite complex, its transfer performance as a source domain is not particularly good. This is mainly because SSV2 videos primarily focus on the temporal dimension, whereas other datasets are more sensitive to the spatial dimension. Therefore, we can conclude that the choice of the source domain is quite important in cross-domain few-shot action recognition.
Additionally, in \Cref{table:sandt}, we observe that when the target domain is fixed and the model is trained with different source domains, there is an overall significant drop in final accuracy compared to training with in-domain datasets.

\begin{table*}[htbp]
  \centering
  \caption{Study the results of our model according to different source datasets. Each experiment is consistent with the paper, involving both pre-training and meta-training stages.
  The Kinetics in the table is Kinetics100-Small.}
    \begin{tabular}{l|cc|cc|cc|cc|cc|cc}
    \toprule
    \multirow{2}[2]{*}{Source to Target} & \multicolumn{2}{c|}{Kinetics} & \multicolumn{2}{c|}{HMDB51} & \multicolumn{2}{c|}{UCF101} & \multicolumn{2}{c|}{SSV2} & \multicolumn{2}{c|}{Diving48} & \multicolumn{2}{c}{RareAction} \\
          & \multicolumn{1}{l}{1-shot} & \multicolumn{1}{l|}{5-shot} & \multicolumn{1}{l}{1-shot} & \multicolumn{1}{l|}{5-shot} & \multicolumn{1}{l}{1-shot} & \multicolumn{1}{l|}{5-shot} & \multicolumn{1}{l}{1-shot} & \multicolumn{1}{l|}{5-shot} & \multicolumn{1}{l}{1-shot} & \multicolumn{1}{l|}{5-shot} & \multicolumn{1}{l}{1-shot} & \multicolumn{1}{l}{5-shot} \\
    \midrule
    Kinetics   &47.08& 63.21 & 39.62  & 54.90 &  65.0 &81.90 & 26.78  &32.10 & 34.20  &42.28  &37.50 &53.3\\
    HMDB51     &36.94  & 48.58  &41.40& 55.88  &46.86 & 66.9& 23.4  & 28.0 & 32.76 &42.54&33.62 &46.0 \\
    UCF101     & 42.26 & 57.06 &37.70&52.92 & 59.66 &79.13 &22.66 & 29.12  &32.20 & 41.12  & 34.82 & 50.14 \\    
    SSV2       & 40.58 &53.20  &33.64 &45.74 & 49.82& 67.62& 31.72& 44.14& 29.32 & 37.42 &33.74 & 46.84 \\
    Diving48   &31.38&39.24&27.94&35.14&39.70&53.46 &22.46 &24.56 &36.52 & 46.08&30.04& 42.52 \\
    RareAction & 34.28& 44.84 &29.70 & 40.29 &41.64& 59.36& 23.14 & 26.82& 34.22 &44.20 &66.72& 85.70\\
    \bottomrule
    \end{tabular}%
  \label{table:sandt}%
\end{table*}%

\subsection{Generalization for the feature of attention}
\begin{figure}[htbp]
    \centering
    \subfloat[Attention of UCF101, the category is ``Haircut''. ]{
    \label{fig:6.a}
    \includegraphics[width=8cm]{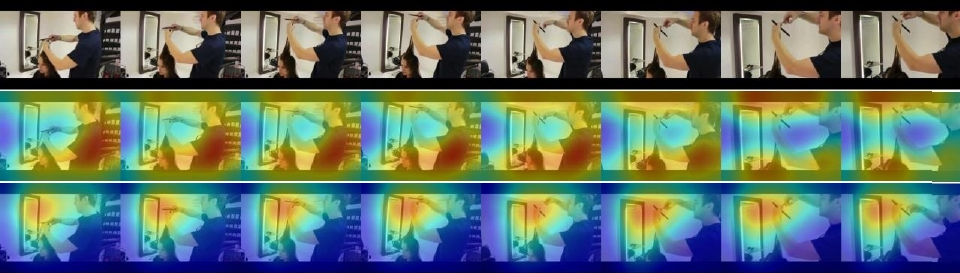}
    }\\
    \subfloat[Attention of HMDB51, the category is ``dribble''.]{
    \label{fig:6.b}
    \includegraphics[width=8cm]{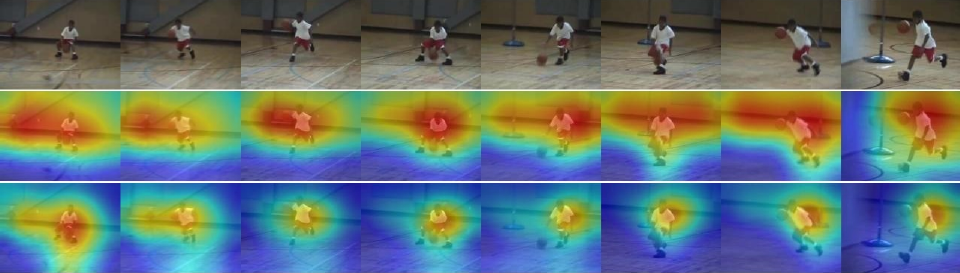}
    }\\
    \subfloat[Attention of SSV2, the category is ``Pretending to take something out of something''.]{
    \label{fig:6.c}
    \includegraphics[width=8cm]{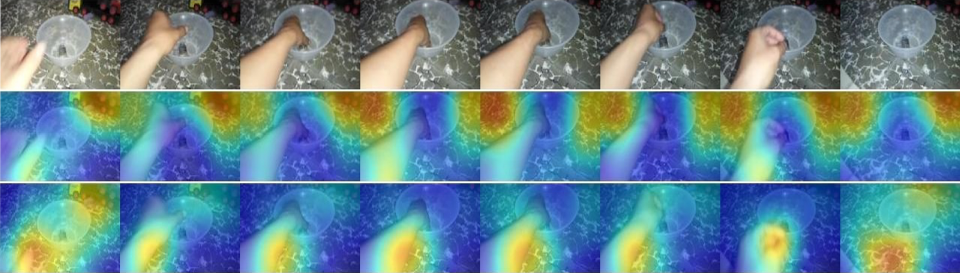}
    }\\
    \subfloat[Attention of rareAction, the category is ``cut keyboard''.]{
    \label{fig:6.d}
    \includegraphics[width=8cm]{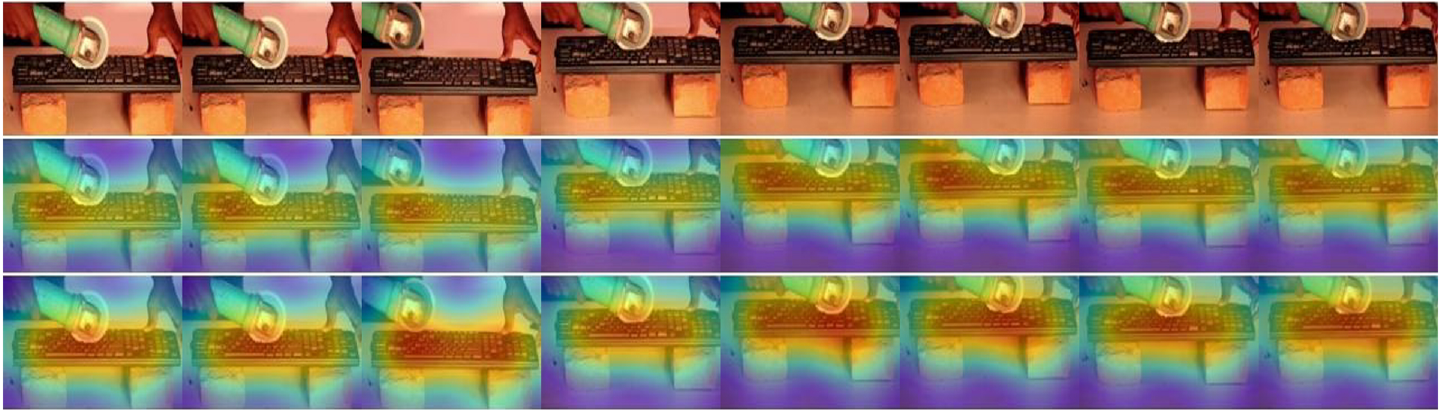}
    }\\
    \subfloat[Attention of Diving48, the category is the No.31 Diving Routine: ``Inward, 15som, NoTwis, PIKE''.]{
    \label{fig:6.e}
    \includegraphics[width=8cm]{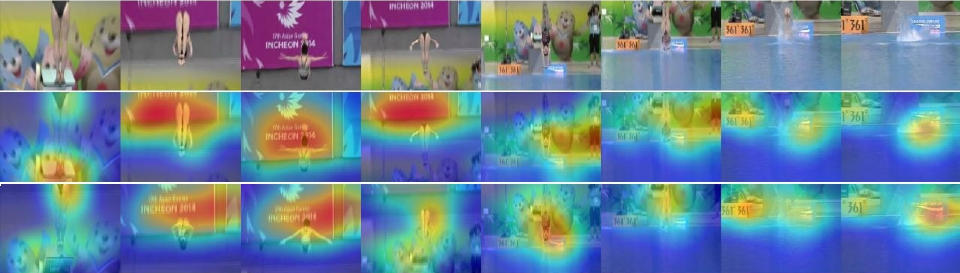}
    }
    \caption{
     Grad-CAM visualization for the STRM and our DMSD. We provide attention maps representing distinct target domains for the UCF101, HMDB51, SSV2, Diving48, and RareAct. These maps are generated from the backbone of the model trained on the Kinetics datasets. 
    }
 \label{fig:6}
\end{figure}

We utilize CAM \cite{selvaraju2017grad-cam} to visualize the features of each frame and verify that our method can learn features that contain rich semantic information conducive to generalization. The results are presented \Cref{fig:6}. 
For each subplot, the first row presents the raw frames, the second row illustrates the attention on the STRM, and the third row showcases the attention on our model's backbone. 
Cref{fig:6.a}, for UCF101, it is evident that our model's attention is predominantly focused on objects associated with cutting hair. 
\Cref{fig:6.b}, for HMDB51, our model demonstrates the ability to disregard extraneous background information. 
\Cref{fig:6.c}, for SSV2, elucidates that our model's attention is more precisely attuned to relevant objects compared to the attention exhibited by STRM. STRM focuses more on ``something'' (perhaps small balls) and unrelated backgrounds.
\Cref{fig:6.d}, for rareAction, our model's attention is primarily directed towards the objects related to the action of the ``cutting keyboard'', just as cutting gear and part of the keyboard near the gear. 
However, the STRM pays less attention to this area.
In \Cref{fig:6.e}, our model's attention exhibits greater precision in certain frames related to each sub-action when compared to STRM based on the Diving dataset. 
Overall, in the cross-domain setting where training is conducted on the source domain and inference is performed on the target domain, Our DMSD's attention mechanism exhibits greater precision.

\subsection{Distribution comparison}
\begin{figure*}[htbp]
    \centering
    \subfloat[Baseline]{
    \label{fig:7.a}
    \includegraphics[width=3.5cm]{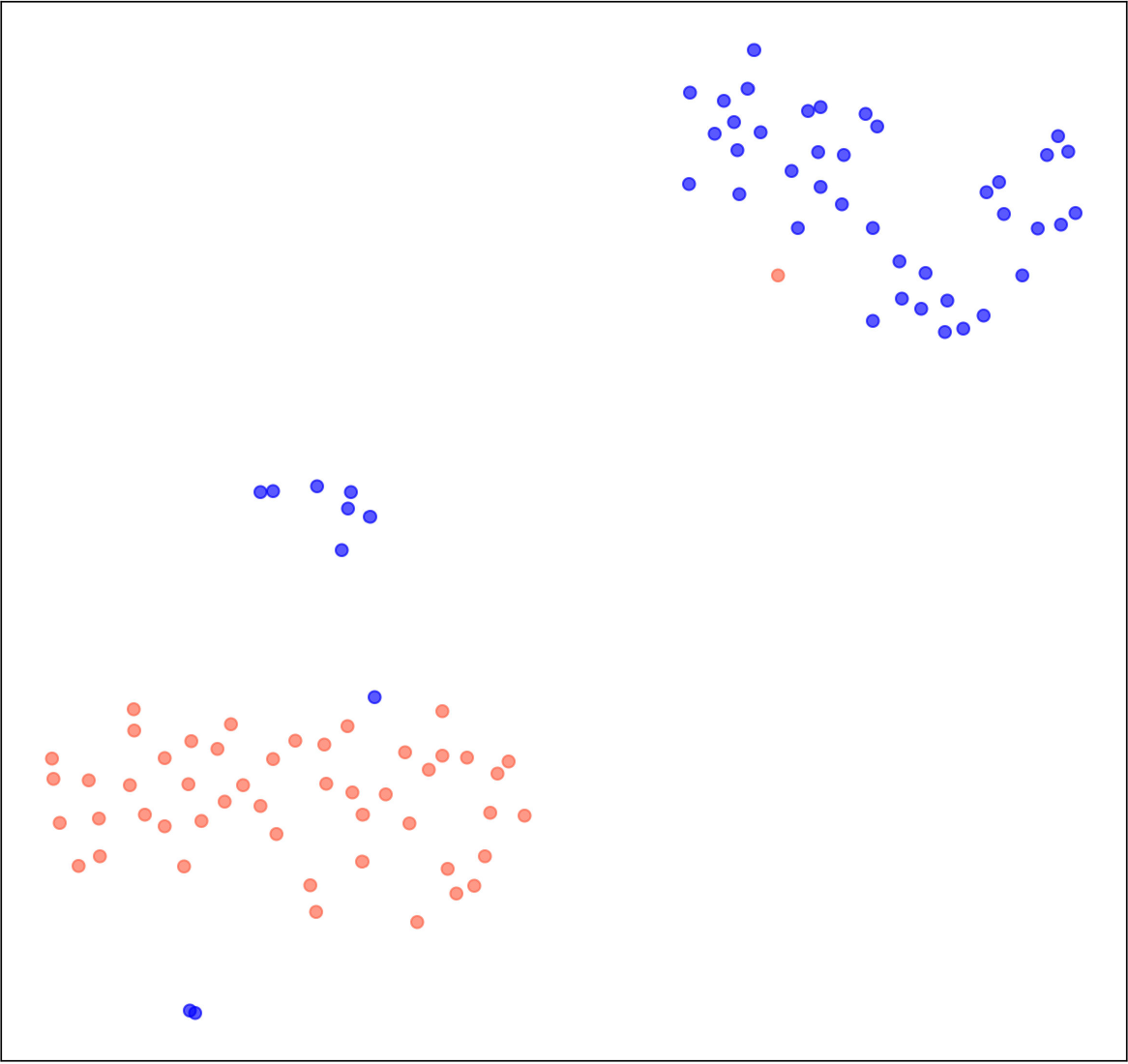}
    }
    \subfloat[SDME]{
    \label{fig:7.b}
    \includegraphics[width=3.5cm]{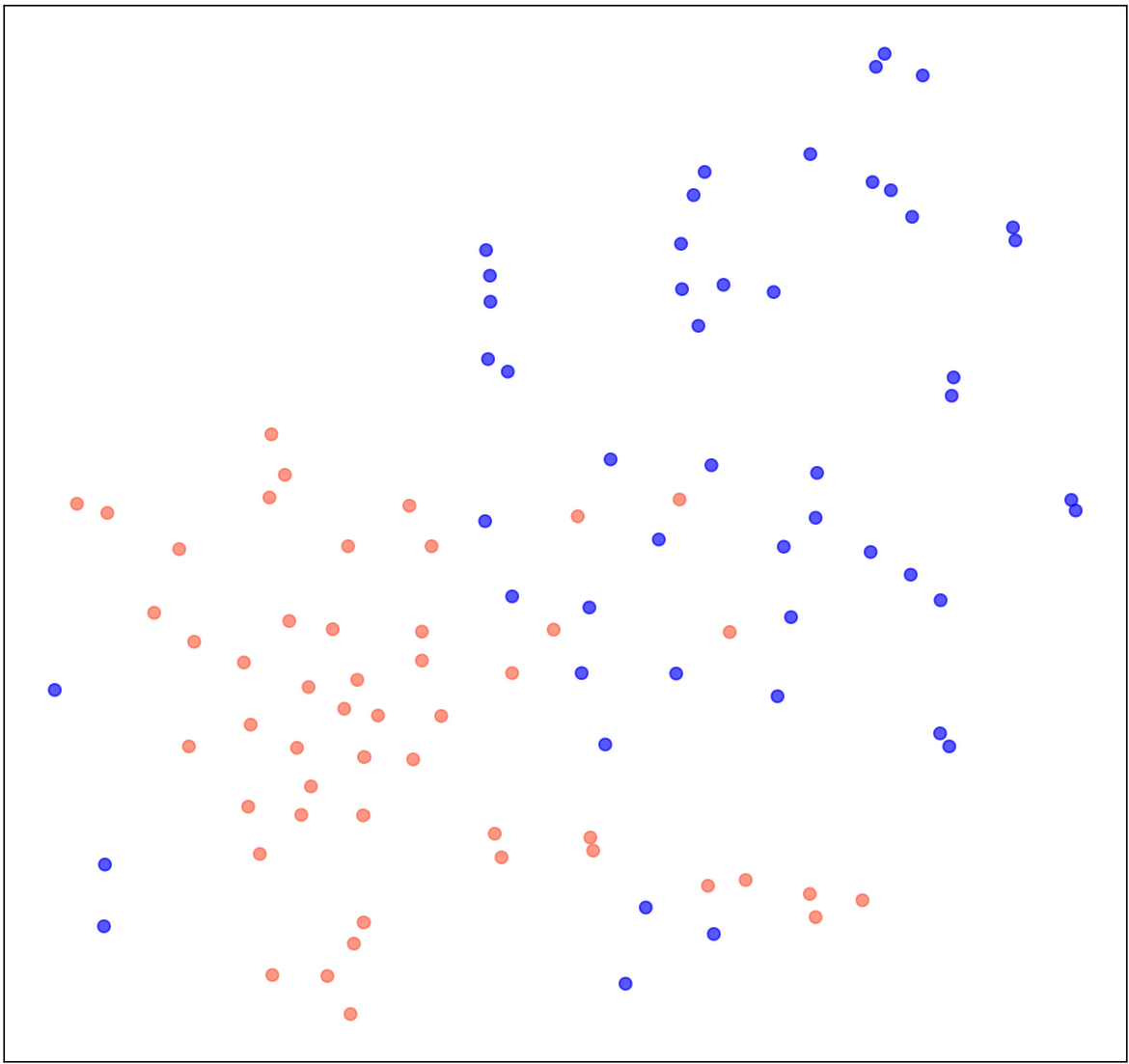}
    }
    \subfloat[SDTE (no distillation)]{
    \label{fig:7.c}
    \includegraphics[width=3.5cm]{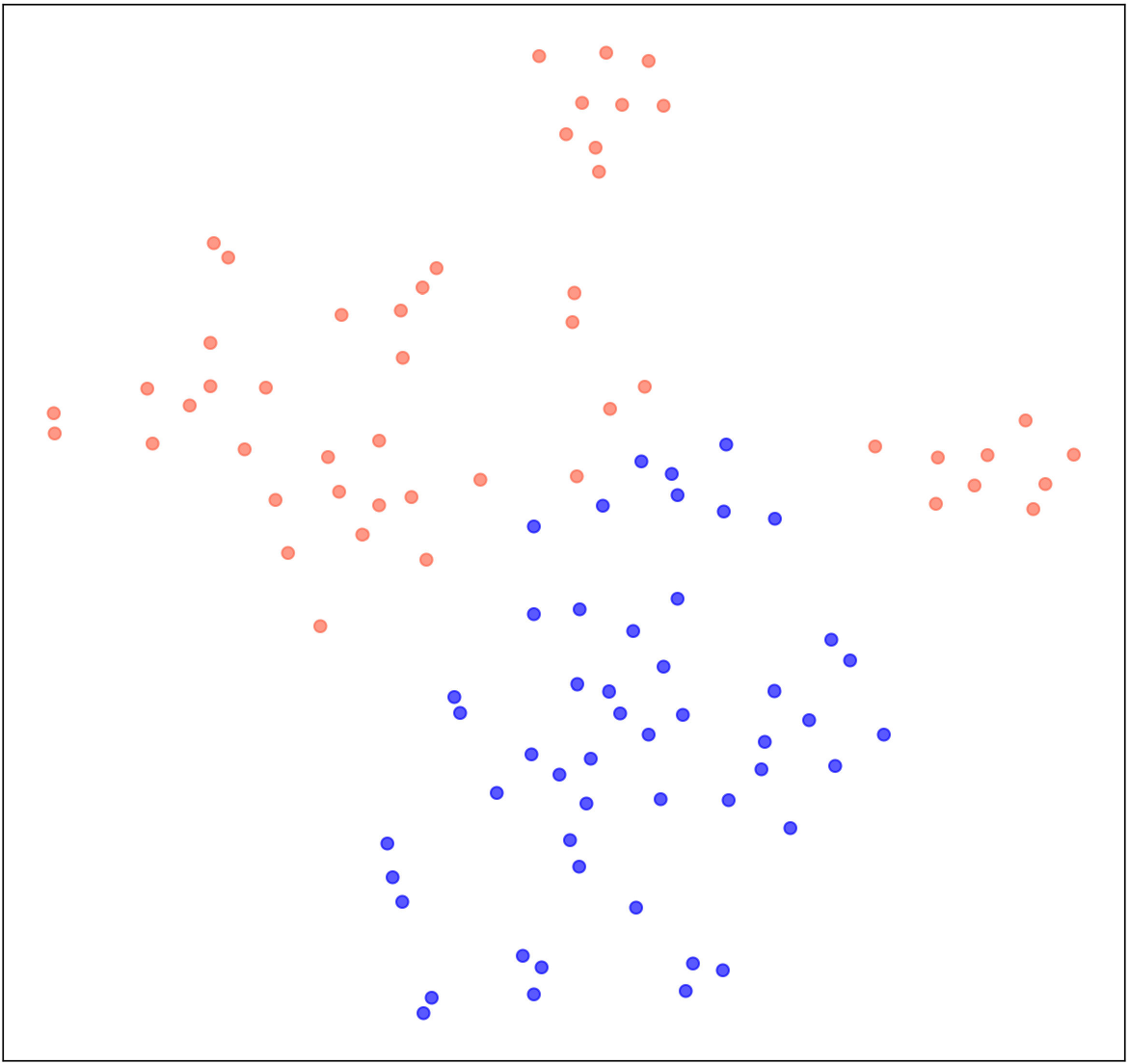}
    }
    \subfloat[SDTE (with distillation)]{
    \label{fig:7.d}
    \includegraphics[width=3.5cm]{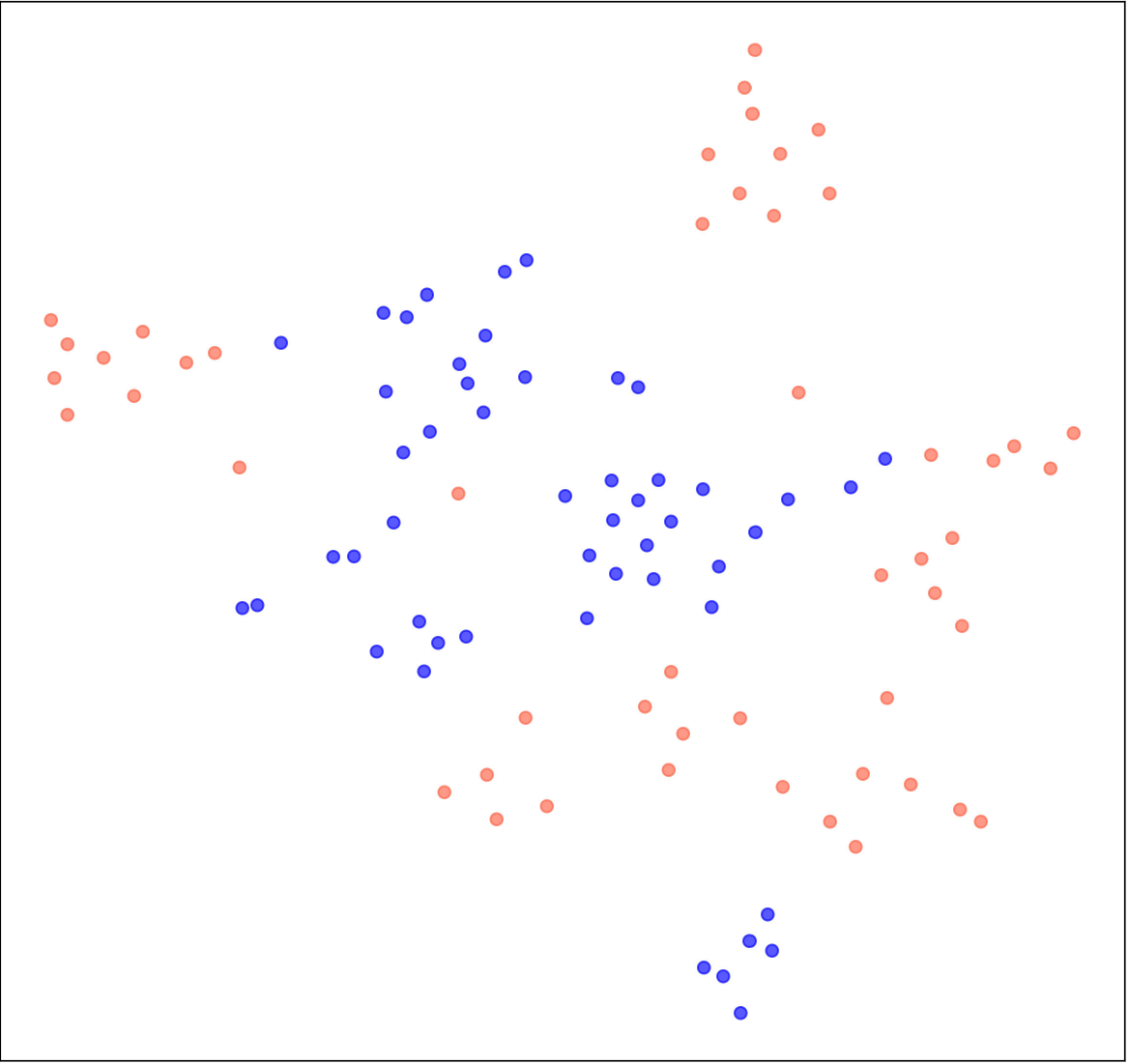}
    }
    \caption{The visualization for distribution of the target domain Diving48 performed with t-SNE. We use the Kinetics100-Small as the source domain and Diving48 as the target domain.
    (a) Distribution of source and target features from the backbone. 
    (b) Distribution of source and target features from the SDME in the source-mixed branch. 
    (c) Distribution of source and target features from the SDTE in the original-source branch (using no distillation from source-mixed branch). 
    (d) Distribution of source and target features from the SDTE in the original-source branch (using distillation from source-mixed branch).
    }
 \label{fig:7}
\end{figure*}

\begin{figure*}[htbp]
    \centering
    \subfloat[Baseline]{
    \label{fig:8.a}
    \includegraphics[width=3.5cm]{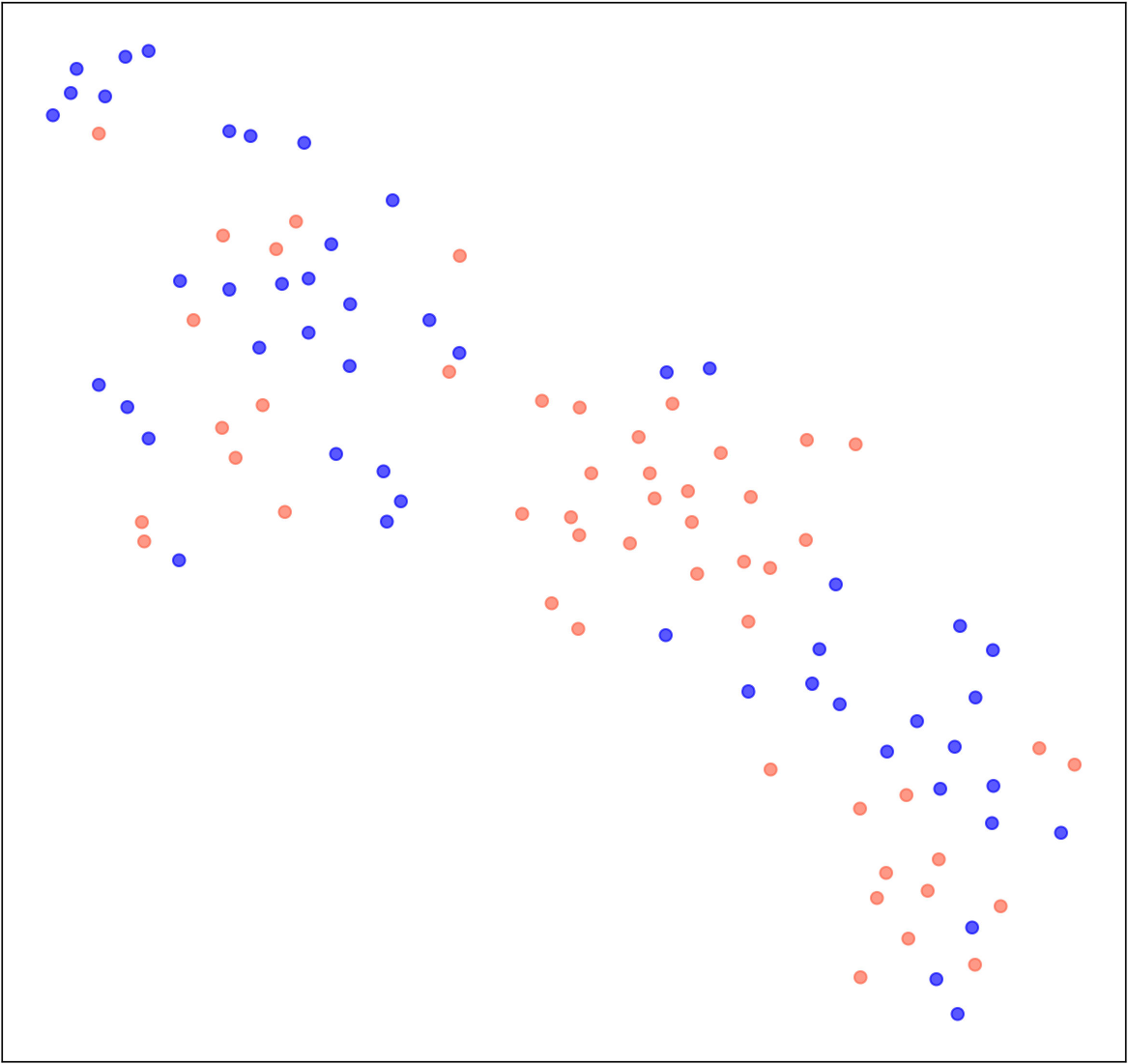}
    }
    \subfloat[SDME]{
    \label{fig:8.b}
    \includegraphics[width=3.5cm]{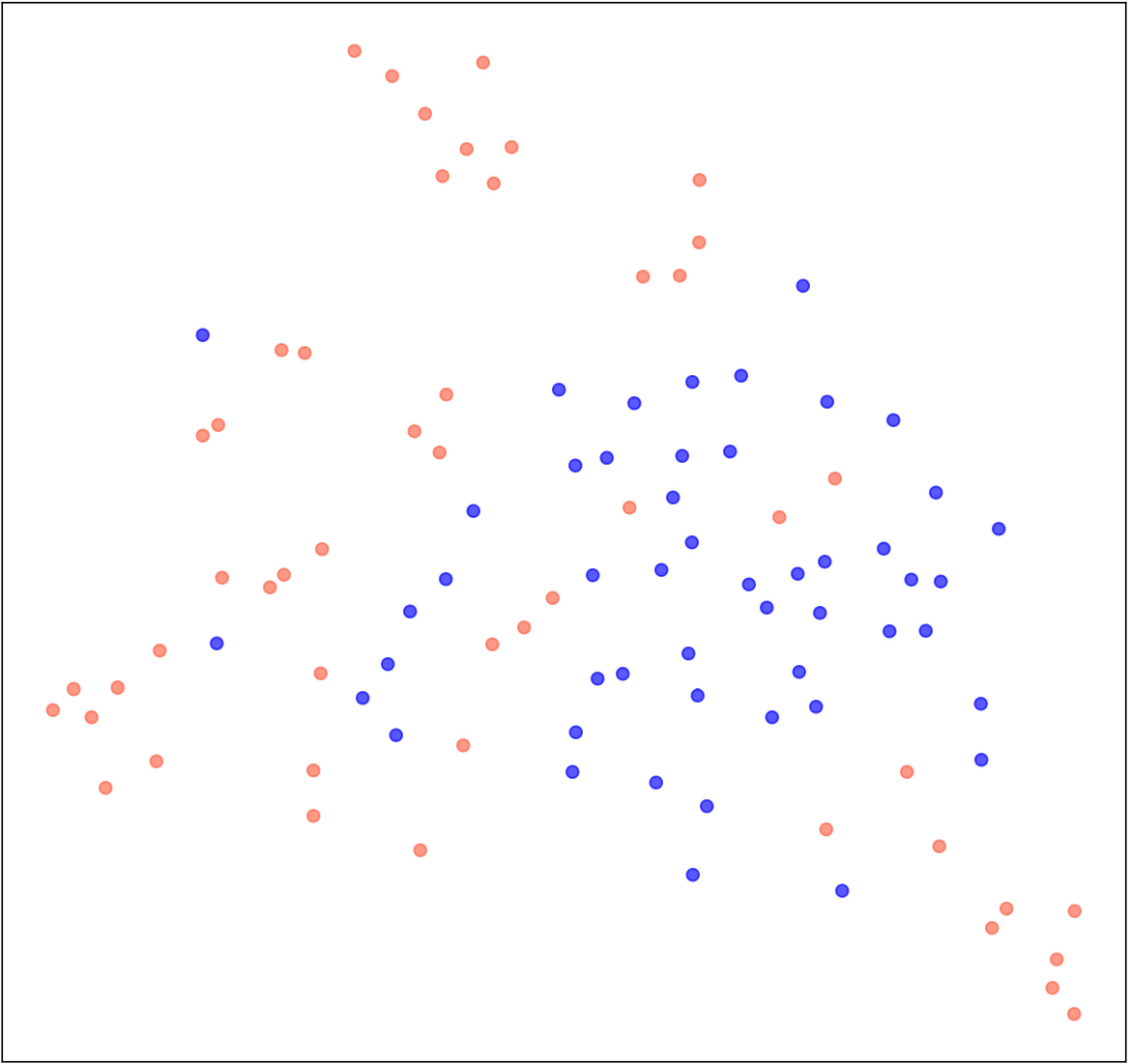}
    }
    \subfloat[SDTE (no distillation)]{
    \label{fig:8.c}
    \includegraphics[width=3.5cm]{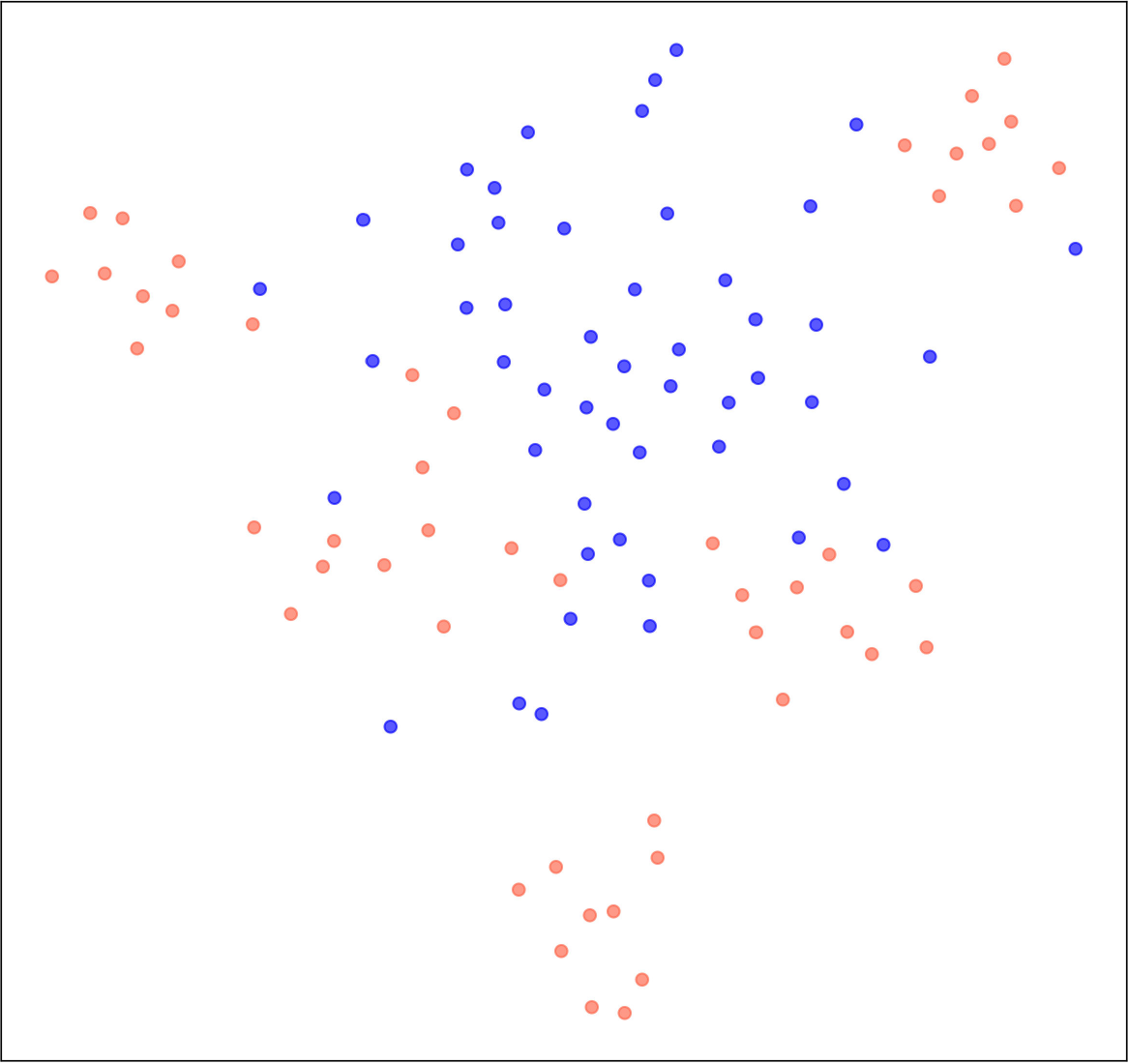}
    }
    \subfloat[SDTE (with distillation)]{
    \label{fig:8.d}
    \includegraphics[width=3.5cm]{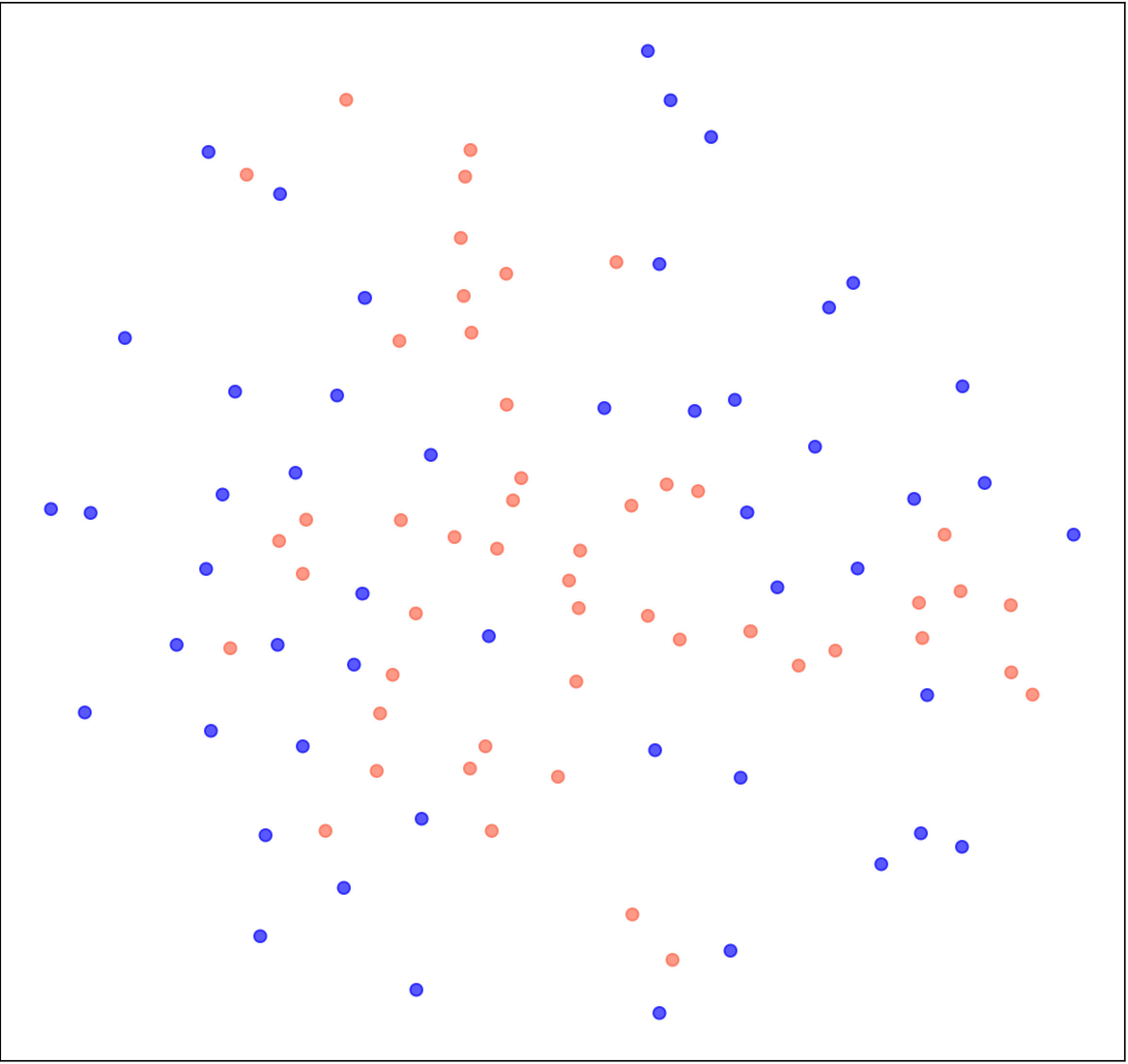}
    }
    \caption{The visualization for distribution of the target domain SSV2 is performed with t-SNE. We use the Kinetics100-Small as the source domain and Diving48 as the target domain. The annotation of subplots is similar to \Cref{fig:7}.
    }
 \label{fig:8}
\end{figure*}

To demonstrate the enhancement of target domain feature distribution achieved by training our model on the Kinetics, we visualize the feature distributions learned by the different parts of DMSD in Figure \ref{fig:7} and \ref{fig:8} using t-SNE \cite{van2008visualizing-tsne} on SSV2 and Diving48. The blue points are source domain data, and the red are target domain data. 
The \Cref{fig:7.a} and \Cref{fig:8.a} illustrate the feature distribution of both the source and target domains through the backbone. 
The \Cref{fig:7.b} and \Cref{fig:8.b} display feature distributions following the Domain Mixer Encoder processing in the mixed-source branch. 
The \Cref{fig:7.c} and \Cref{fig:8.c} show feature distributions following the Domain Temporal Encoder in the original-source branch when trained without distillation from the Mixed-source branch.
The \Cref{fig:7.d} and \Cref{fig:8.d} present feature distributions of the Domain Temporal Encoder in the original-source branch after applying a distillation with the Domain Mixer Encoder serving as the teacher. 
From the figures, the following observations can be made:
(1) The feature distributions of the source and target domains exhibit significant differences after backbone feature extraction.
(2) After processing through the Domain Mixer Encoder, the feature distributions still show some differences, although the distance between samples from different domains is reduced.
(3) Since the Domain Temporal Encoder extracts features and reconstructs data for the target domain, the feature distributions of the source and target domains begin to merge when only the original-source branch is used.
(4) After distilling knowledge from the mixed-source branch, the feature distributions of the source and target domains processed by the Domain Temporal Encoder in the original-source branch show better integration.
These visual analyses demonstrate the effectiveness of our Encoder-Decoder mechanism and Knowledge Distillation approach for cross-domain few-shot action recognition.

\begin{figure}[htbp]
    \centering
    \subfloat[Distribution of HMDB51]{
    \label{fig:9.a}
    \includegraphics[width=7cm]{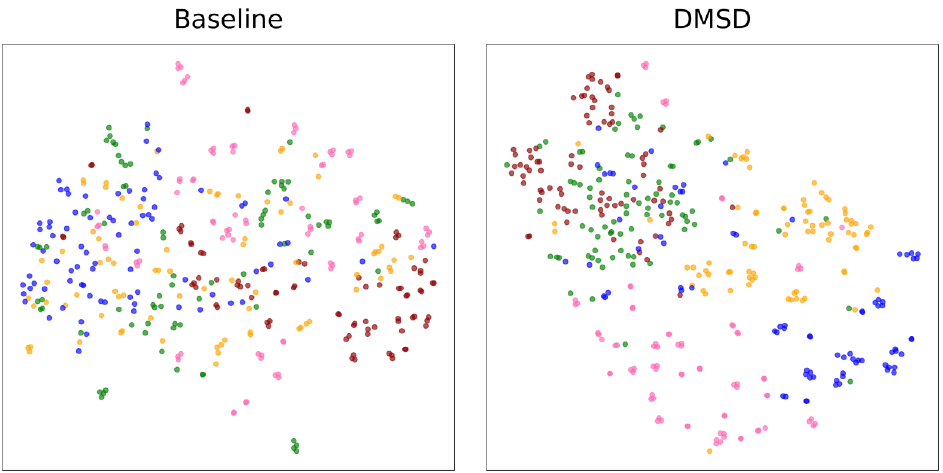}
    }\\
    \subfloat[Distribution of UCF101]{
    \label{fig:9.b}
    \includegraphics[width=7cm]{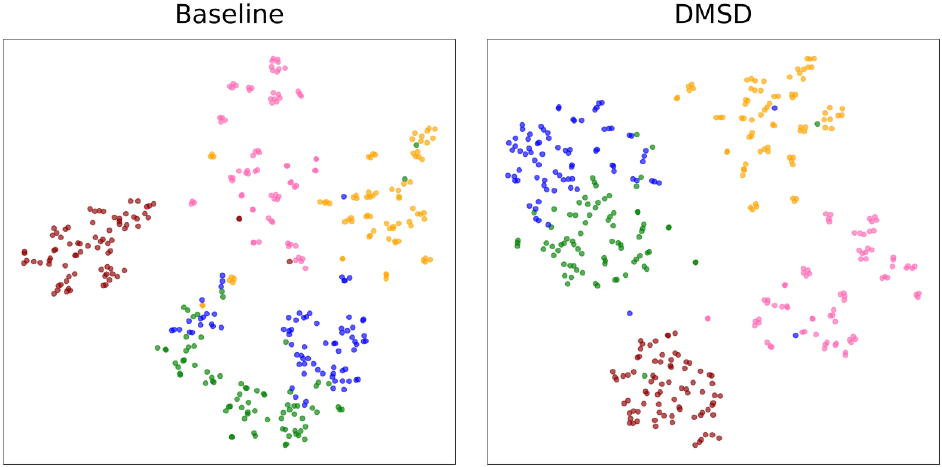}
    }
    \caption{The visualization of feature distribution in each category of the target domain. It is performed with t-SNE. 
    (a) Distribution of target domain HMDB51 based on the source domain of Kinetics100-Small.
    (b) Distribution of target domain UCF101 based on the source domain of Kinetics100-Small.
    }
 \label{fig:9}
\end{figure}

To further validate the transferability of our proposed method, we present the feature visualization results for the meta-test of UCF101 and HMDB51 in \Cref{fig:9}, selecting five categories of samples from each dataset. Compared to the baseline method, DMSD demonstrates superior intra-class cohesion and more pronounced inter-class separation in the feature clustering structure. These results indicate that by incorporating unlabeled target samples through Cycle Consistency Loss and performing knowledge distillation on the mixed-source branch mixed with the target domain, our method effectively reduces the domain gap and enhances the model's generalization capabilities.

\section{Conclusion}
In this paper, we propose a methodology called Distillation from Mixed-Source Domain for CDFSAR. The core idea is to fully utilize source and target domain videos and overcome overfitting under a fine-tuning paradigm when only a few labeled target domain samples are available.
We combine meta-learning and supervised learning. 
On the one hand, we aim to learn more transferable knowledge through full-label supervised learning. On the other hand, we smoothly transition meta-learning from the training stage to the testing stage, thereby avoiding fine-tuning.
In our approach, the mixed-source branch fuses target domain information, while the original-source branch classifies samples and extracts general features from both source and target domains. 
Finally, we use distillation to encourage the original-source branch to learn from the knowledge distilled from the mixed-source branch.
In the future, exploring scientific methods for selecting and merging target features is worth investigating. Additionally, we plan to explore using some labeled target domain features.

\section{Acknowledgements}
This work is supported by 
the National Key Research and Development Program(Grant No. 2019YFB2102500) and
the Guangdong Basic and Applied Basic Research Foundation (Grant No. 2021A1515011913).

\bibliographystyle{cas-model2-names}
\bibliography{dmsd_main}
\end{document}